\documentclass{article} 
\usepackage[preprint]{colm2025_conference} \usepackage{microtype}
\usepackage{hyperref}
\usepackage{url}
\usepackage{booktabs}

\usepackage{lineno}

\definecolor{darkblue}{rgb}{0, 0, 0.5}
\hypersetup{colorlinks=true, citecolor=darkblue, linkcolor=darkblue, urlcolor=darkblue}

\usepackage[utf8]{inputenc} 
\usepackage[T1]{fontenc}    
\usepackage{hyperref}       
\usepackage{url}            
\usepackage{booktabs}       
\usepackage{amsfonts}       
\usepackage{nicefrac}       
\usepackage{microtype}      
\usepackage{xcolor}         

\usepackage{amsmath} 
\usepackage{cleveref}
\usepackage{subcaption}
\usepackage{graphicx}
\usepackage{booktabs}
\usepackage{threeparttable}
\usepackage{longtable}
\usepackage{array}
\usepackage{wrapfig}
\usepackage{multirow}
\usepackage[most]{tcolorbox}
\tcbuselibrary{skins}
\usepackage{multicol}
\usepackage{adjustbox}
\usepackage{colortbl}
\usepackage{xspace}

\crefname{figure}{Fig.}{Figs.}
\Crefname{figure}{Fig.}{Figs.}
\crefname{appendix}{Appx.}{Appx.}
\crefname{table}{Tab.}{Tables}
\Crefname{table}{Tab.}{Tables}
\crefname{section}{Sec.}{Sec.}
\Crefname{section}{Sec.}{Sec.}
\crefname{equation}{Eq.}{Eqs.}
\Crefname{equation}{Eq.}{Eqs.}
\crefformat{equation}{Eq.~#2#1#3}
\Crefformat{equation}{Eq.~#2#1#3}

\newcommand{\titlename}{\textsc{Maye}\xspace}

\title{Rethinking RL Scaling for Vision Language Models: \\A Transparent, From-Scratch Framework and Comprehensive Evaluation Scheme}

\author{
 Yan Ma$^{3,5}$, Steffi Chern$^{5}$, Xuyang Shen$^{2}$, 
 Yiran Zhong$^{2*}$, Pengfei Liu$^{1,4,5}\thanks{\; Corresponding authors. Email: \texttt{zhongyiran@gmail.com,pengfei@sjtu.edu.cn}.}$
 \\
  \\ $^1$Shanghai Jiao Tong University (SJTU)  
  $^2$Minimax\\
  $^3$Fudan University
  $^4$SII
  $^5$Generative Artificial Intelligence Lab (GAIR)
\\
}

\begin{document}

\ifcolmsubmission
\linenumbers
\fi

\maketitle
\begin{abstract}
Reinforcement learning (RL) has recently shown strong potential in improving the reasoning capabilities of large language models and is now being actively extended to vision-language models (VLMs). However, existing RL applications in VLMs often rely on heavily engineered frameworks that hinder reproducibility and accessibility, while lacking standardized evaluation protocols, making it difficult to compare results or interpret training dynamics. This work introduces a transparent, from-scratch framework for RL in VLMs, offering a minimal yet functional four-step pipeline validated across multiple models and datasets. In addition, a standardized evaluation scheme is proposed to assess training dynamics and reflective behaviors. Extensive experiments on visual reasoning tasks uncover key empirical findings: response length is sensitive to random seeds, reflection correlates with output length, and RL consistently outperforms supervised fine-tuning (SFT) in generalization—even with high-quality data. These findings, together with the proposed framework, aim to establish a reproducible baseline and support broader engagement in RL-based VLM research. Code is public and available at: \url{https://github.com/GAIR-NLP/MAYE}.
\end{abstract}

\section{Introduction}
Reinforcement learning (RL) has recently demonstrated remarkable success in enhancing reasoning capabilities of LLMs, particularly on tasks with verifiable answers such as mathematical problem solving~\citep{guo2025deepseekR1,rl_survey}. Inspired by this progress, growing efforts have extended RL to VLMs, aiming to replicate the so-called “R1 moment”~\citep{wang2025multimodal,qwen2025qvqmax}. 
These studies have primarily concentrated on enhancing performance and pushing the state-of-the-art. However, many of these works rely heavily on highly engineered and encapsulated codebases, such as TRL \citep{vonwerra2022trl}, OpenRLHF \citep{hu2024openrlhf}, and verl \citep{sheng2024hybridflow}, making it difficult for newcomers to understand, replicate, or modify the underlying processes. This has led to a gap in the field, particularly for researchers who are not already deeply familiar with both RL and VLMs. As a result, the learning curve for those entering this area remains steep.

We address this gap by introducing a reproducible standard framework for RL in VLMs, which serves as a transparent and accessible foundation for training RL-based VLMs. Unlike prior works that rely on complex, pre-packaged RL libraries, the proposed framework is implemented entirely from scratch, using only standard libraries such as Transformers \citep{wolf-etal-2020-transformers}, FSDP2 \citep{zhao2023pytorch} for distributed training, and vLLM \citep{kwon2023efficient} for inference. This minimal yet functional implementation allows for a clearer understanding of the RL training process and ensures that the core logic is fully transparent, enabling easy customization and experimentation.

By building the framework from the ground up, this work provides a solid foundation for further improvements and extensions in RL for VLMs. It also serves as a crucial resource for beginners, offering a simplified entry point to understanding how RL can be applied to VLMs. This framework, while not aiming to be the most performant or highly optimized, acts as an essential entry into the mechanism of RL in VLMs, much like OpenAI's SpinningUp \citep{SpinningUp2018} for RL, providing significant value to the research community. It can be used both as a base for future RL innovations and as an educational tool for fostering broader engagement with RL-based VLM research.

Besides, while the proposed framework addresses the need for a reproducible RL training process, the evaluation of RL remains a challenging task. Currently, there is no unified or standardized approach to assess RL training in the context of LLMs/VLMs, leaving a significant gap in the field.
To address this, a comprehensive evaluation scheme is introduced, offering a structured framework for assessing RL training effectiveness. Unlike instruction-tuning~\citep{zhang2023instruction} or DPO \citep{rafailov2023DPO}, where a single performance score is often deemed sufficient, RL training involves dynamic, fluctuating performance that is sensitive to several factors such as initialization and random seed variation~\citep{henderson2018deep,andrychowicz2020matters}. Reporting a single final score can overfit to incidental fluctuations, compromising the reproducibility and generalization of results.
The proposed evaluation scheme, detailed in~\cref{sec:evaluation_scheme}, emphasizes capturing the training dynamics across multiple stages. Key performance metrics include accuracy curves under different generation settings, as well as behavioral indicators such as response length and reflection ratio. By incorporating fine-grained reflective behavior metrics, the scheme ensures a more nuanced and transparent evaluation of RL’s effectiveness.

Based on the proposed framework, RL experiments are conducted on multiple VLMs across diverse visual reasoning datasets. Each experiment is independently repeated to account for training variance and ensure reproducibility—consistent with best practices in the RL community~\citep{colas2018many,agarwal2021deep}. By applying the evaluation scheme, several notable findings emerge: response length is highly sensitive to random seeds; reflective behaviors strongly correlate with length dynamics; and RL consistently demonstrates superior generalization compared to SFT, even when the latter is trained with high-quality supervision. These findings are detailed in~\cref{sec:experiment}.

In this work, three core contributions are made: 1) A reproducible and from-scratch RL framework for VLMs.  
A transparent four-step pipeline is implemented without relying on existing RL toolkits, validated across multiple VLMs and datasets.
2) A standardized evaluation scheme tailored for RL training.  
The scheme captures training dynamics and reflective behavior, offering robust and reproducible benchmarks for future studies.
3) Empirical insights into length, reflection, and generalization.  
Analysis reveals the coupling between reflection and response length, and highlights RL’s superior generalization over SFT, even with high-quality supervision.

\section{Preparation}
\label{secappen:preparation}
\begin{wrapfigure}{r}{0.40\textwidth}
    \centering
    \includegraphics[width=0.98\linewidth]{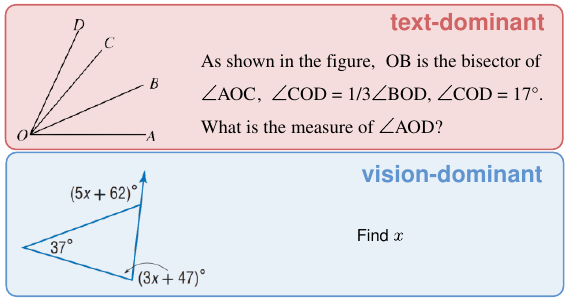}
    \caption{Text-dominant tasks rely on text with visual support; vision-dominant tasks rely on visuals with textual support.}
    \label{fig:text_vision_dominant}
    \vspace{-3mm}
\end{wrapfigure}

This section outlines the foundational setup required for RL in VLMs. It includes four parts: data, algorithm, reward function, and model. Together, these elements define the training context and ensure that the subsequent RL process proceeds under a coherent and reproducible configuration.
\paragraph{Data}
serves as the foundation for training and evaluation.
Rule-based RL has demonstrated strong effectiveness in text-based reasoning tasks where answers can be explicitly verified~\citep{guo2025deepseekR1, rl_survey}.
In this report, we continue to focus on verifiable mathematical reasoning problems to construct training and evaluation queries.
To account for the varying granularity of information provided by these two modalities, we categorize visual mathematical reasoning into two subtypes: text-dominant and vision-dominant, as illustrated in~\cref{fig:text_vision_dominant}.
In the text-dominant setting, most of the necessary information is in the text, while the image provides additional support.
In contrast, the vision-dominant setting requires extracting key information directly from the image.
For text-dominant tasks, we use the \textit{mm\_math5k} dataset~\citep{sun2024mmmath}, while for vision-dominant tasks, we use the \textit{geometry3k} dataset~\citep{zheng2025easyr1}.
The partitioning of training, validation, and test sets for both datasets is detailed in~\cref{tab:dataset_stats}.
To assess the out-of-distribution generalization of RL in VLMs, we construct the test set for \textit{mm\_math5k} using 100 problems sampled from MathVerse~\citep{zhang2024mathverse}.
Additionally, to prevent reward hacking, all problems are designed as numerical computation tasks, ensuring that RL-based models focus on reasoning rather than exploiting spurious correlations in reward signals \citep{kimi2025k1.5}.
\begin{table}[tb!]
  \centering
  \resizebox{0.98\textwidth}{!}{ 
  \begin{threeparttable}
    \caption{Dataset Statistics, $^\dagger$ means that samples are from the MathVerse benchmark.}
    \label{tab:dataset_stats}
    \begin{tabular}{ccccc>{\raggedright\arraybackslash}p{4cm}}
      \toprule
      Dataset Name & Training Set Size & Validation Set Size & Test Set Size & Task & \multicolumn{1}{c}{Data Source} \\
      \midrule
      mm\_math5k & 5000 & 100 & ~100\textsuperscript{$\dagger$} & Text-dominant & \raisebox{-0.2\height}{\includegraphics[height=1.2em]{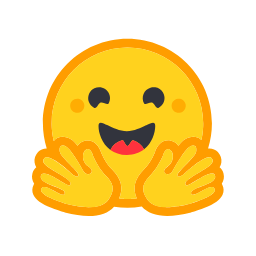}}THU-KEG/MM\_Math \\
      geometry3k & 2101 & 300 & 601 & Vision-dominant & \raisebox{-0.2\height}{\includegraphics[height=1.2em]{figures/hf.png}}hiyouga/geometry3k \\
      \bottomrule
    \end{tabular}
    \vspace{-3mm}
  \end{threeparttable}
  }
\end{table}
\paragraph{Algorithm} selection plays a crucial role in RL for VLMs.
Policy-based RL, particularly methods that discard value functions, has become the mainstream approach.
Among them, Group Relative Policy Optimization (GRPO) \citep{shao2024deepseekmath} has been the most widely used in recent research.
In this report, we explore an alternative approach, Reinforce++ \citep{hu2025reinforce++}, to investigate its potential as another option for RL in VLMs and assess its effectiveness in VLM training.
Following~\cite{xie2025logicrlunleashingllmreasoning}, we also incorporate a KL divergence penalty between the policy and the reference model, which introduces an additional loss term.
The modified update objective is given by:
{\small
\begin{align}
\label{eq:ppo_loss}
&\mathcal{L}^{\text{CLIP}}(\theta) = \mathbb{E}_{[q \sim P(q), o_q \sim \pi_{\theta_{\text{old}}}(o|q)]} \\
&\frac{1}{|o_q|}\sum_{t=1}^{|o_q|}\left\{
\min \left[
\frac{\pi_\theta(o_{q,t}|q,o_{q,<t})}{\pi_{\theta_{\text{old}}}(o_{q,t}|q,o_{q,<t})}\hat{A}_{t},\, \text{clip}\left(\frac{\pi_\theta(o_{q,t}|q,o_{q,<t})}{\pi_{\theta_{\text{old}}}(o_{q,t}|q,o_{q,<t})}, 1-\epsilon, 1+\epsilon\right)\hat{A}_{t} \right] - \beta_{\text{loss}} \mathbb{D}_{\text{KL}}\left[\pi_\theta \| \pi_{\text{ref}}\right]
\right\} \nonumber \\
& \text{Where as}~~~\hat{A}_t = \sum_{k=t}^{|o_q|} \gamma^{k-t}\left\{
\underbrace{\text{\textbf{I}}(o_{q,t}=\left[\text{EOS}\right])r(q, o_q)}_{\text{Rule-based reward}} - \beta_{\text{rew}}\underbrace{\mathbb{D}_{\text{KL}}\left[\pi_\theta(o_{q,t}|q,o_{q,<t}) \| \pi_{\text{ref}}(o_{q,t}|q,o_{q,<t})\right]}_{\text{Token-level KL reward}}\right\} \nonumber
\end{align}
}
$P(q)$ represents the distribution of queries, and $o_q$ denotes the sequence of response tokens. $\epsilon$ constrains the probability ratio $\frac{\pi_\theta(a_t|s_t)}{\pi_{\theta_{\text{old}}}(a_t|s_t)}$ within $\left[1-\epsilon, 1+\epsilon\right]$.
$\hat{A}_t$ represents the estimated advantage for token $t$, which plays a crucial role in determining the direction of parameter updates.
The discount factor $\gamma \in [0,1]$ is fixed to $1$ in our experiments. The identity function $\text{\textbf{I}}(o_{q,t}=\left[\text{EOS}\right])$ evaluates to 1 when the $\text{<EOS>}$ token is reached, and 0 otherwise.  
$\mathbb{D}_{\text{KL}}$ follows the k3 formulation~\citep{schulman2025klapprox}, which provides an unbiased estimation. Additionally, $\beta_{\text{rew}}$ is the coefficient for the KL reward, while $\beta_{\text{loss}}$ is the coefficient for KL penalty loss.
It is important to note that in the subsequent experiments, we only applied the KL penalty loss while discarding the KL reward by setting $\beta_{\text{rew}}$ to 0.
Modifications to the algorithm remain consistent across all experiments.

\paragraph{Reward Function} serves as a rule-based signal for guiding the RL training process. A correct final answer receives a reward of +1; otherwise, 0. A secondary language reward penalizes responses containing non-English characters to discourage multilingual drift. Format rewards are deliberately omitted to avoid constraining the model’s output patterns during learning~\citep{zeng2025simplerlzooinvestigatingtamingzero}.

\paragraph{Model} capability determines whether its cognitive abilities, such as verification and reflection, can be effectively activated.
We choose Qwen-VL series for two key reasons.
First, based on the findings of~\citep{gandhi2025cognitive} and the prevailing choices in the research community, these models have demonstrated strong potential for test-time scaling.
Second, they are natively integrated into Transformers \citep{wolf-etal-2020-transformers}, making them highly accessible and convenient to use.
Therefore, we select Qwen2/2.5-VL-Instruct \citep{wang2024Qwen2-VL,bai2025Qwen2.5-VL} as our backbone models.
\section{\titlename Framework: A Transparent, From-Scratch RL Framework for VLM}
\label{sec:methodology}
\begin{figure}[tb]
    \centering
    \includegraphics[width=0.98\textwidth]{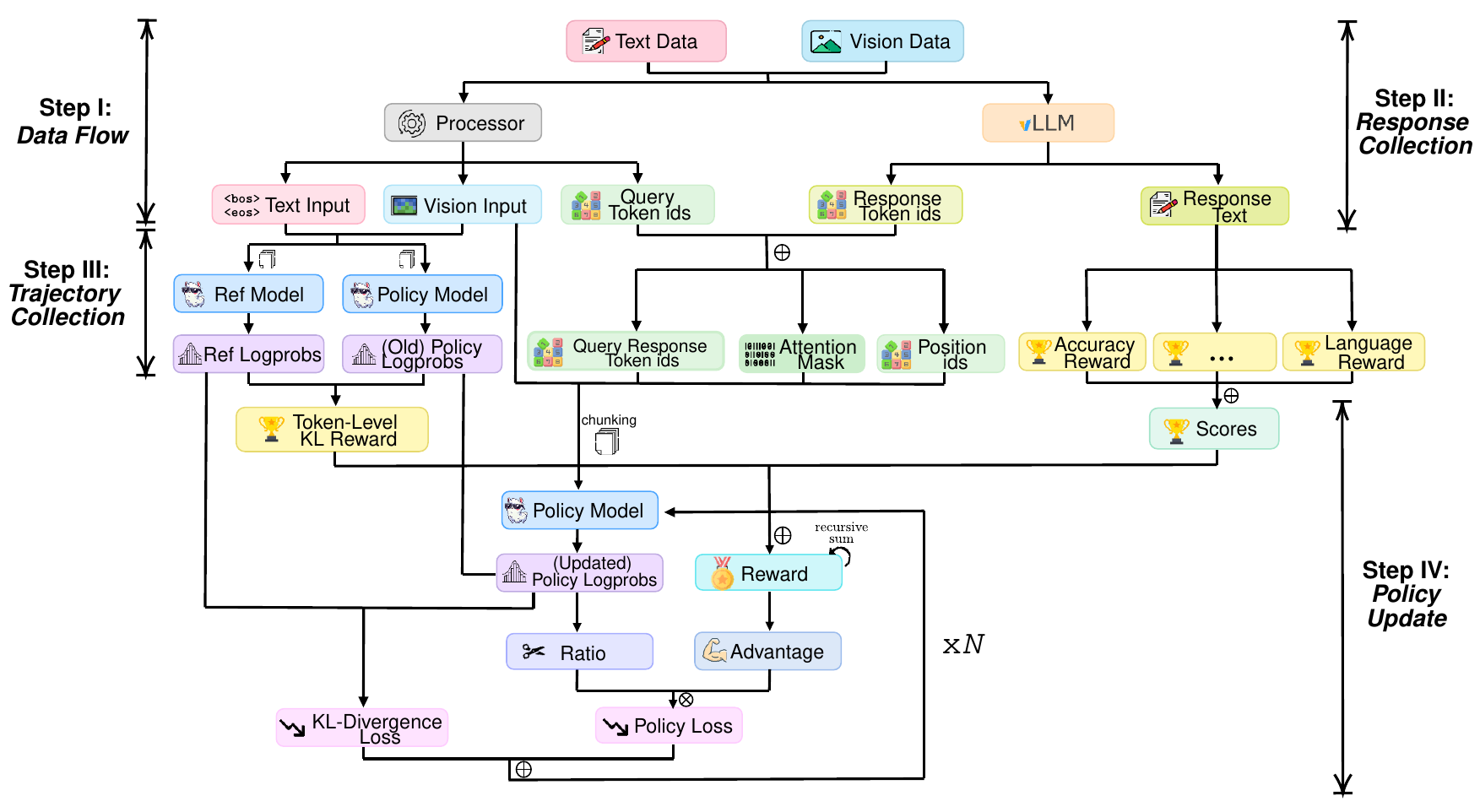}
    \caption{Overview of \titlename framework. The process is divided into four steps. Each step integrates various components, including text and vision data, policy models, and reward signals.}
    \label{fig:framework}
\end{figure}

This section presents the \textbf{\titlename framework}, a transparent, from-scratch RL training pipeline for VLMs, designed as a \textit{reproducible and standardized baseline}. Rather than introducing yet another training system, the framework distills RL into four components—\textit{data flow}, \textit{response collection}, \textit{trajectory generation}, and \textit{policy update}—each made explicit and modular.
\paragraph{Setup}
From a high-level perspective, Hydra \citep{Yadan2019Hydra} is used to manage experiment configurations, Transformers~\citep{wolf-etal-2020-transformers} for modeling VLMs, FSDP2~\citep{zhao2023pytorch} for distributed training, and vLLM \citep{kwon2023efficient} for collecting responses for multimodal queries.
Training and inference are conducted on separate GPU devices.  
Before training begins, the system loads configurations from a YAML file and then initializes the policy and reference models, dataloaders for the training, validation, and test sets, the optimizer, training parameters, the learning rate scheduler, and the vLLM engine.

It is worth noting that VLMs typically consist of a ViT encoder, an MLP connector, and a LLM backend.
Thus, selecting which components to freeze or train is crucial.
Based on preliminary experiments, training the connector and ViT on several thousands samples does not yield significant performance improvements but slows down training speed.
Since the lower layers of the LLM also participate in processing visual inputs~\citep{zhu2025remedy}, there is no concern that the model's visual capabilities will be left untuned. Therefore, all experiments solely train the LLM backend.

The RL process involves a variety of parameters and configurations, some of which are easily confused due to overlapping terminology. In particular, commonly used terms such as \textit{batch}, \textit{epoch}, and \textit{step} may refer to different concepts depending on context. \cref{tab:batch_step_terms} provides a concise reference to clarify these definitions.
A complete list of training and hyperparameters is provided in~\cref{sec:parameters}.
For rollout inference, vLLM \citep{kwon2023efficient} is used to accelerate sampling.
To keep the implementation simple, we do not introduce Ray~\citep{moritz2017ray} for managing training or inference task scheduling.
After completing these setup steps, the subsequent implementation follows a four-step iterative process.

\paragraph{Step I: Data Flow}
Under a multimodal setting, each query contains both vision and text data.
As shown in the top-left of~\cref{fig:framework}, the query batch is first processed by a processor provided by Transformers. This step converts raw data into model-compatible inputs, consisting of both textual and visual modalities. The textual input includes token ids sequences—where image slots are padded using special tokens such as \texttt{<image\_pad>}—along with the corresponding attention masks. The visual input is transformed into pixel values and auxiliary features. Additionally, the query token ids from text input will be used to concatenate with the generated response tokens in Step II.

\paragraph{Step II: Response Collection}
This step (top-right of~\cref{fig:framework}) involves collecting responses to queries, which can be accelerated using the inference engine.  
First, the sharded parameters are gathered on the CPU and synchronized to the inference engine.
Then, the processed inputs from all training GPUs are gathered to the inference device, collecting a response for each query, including both response text and token ids. 
After inference, the responses are broadcast back to their corresponding GPUs. Since response lengths vary, padding is applied to ensure an aligned length.

\paragraph{Step III: Trajectory Generation}

A trajectory can be considered as an essential input for model learning.
It is fundamentally a \textit{namedtuple} that contains both the components required for loss computation and the metrics that need to be recorded.

The center of~\cref{fig:framework} illustrates how \textit{text\_input} is updated: the token ids of queries and responses are concatenated, and the corresponding \textit{attention\_masks} and \textit{position\_ids} are recalculated accordingly. These updated inputs are then stored in the trajectory, as they are required to recompute log probabilities during Step IV.
Meanwhile, as illustrated in the middle-left of~\cref{fig:framework}, the (updated) text input and vision input are forwarded through both the policy and reference models to compute log probabilities (logprobs), with the batch being chunked to prevent out-of-memory.
It is important to note that only the logprobs of the response are retained, as RL is a post-training procedure.
Meanwhile, the center of~\cref{fig:framework} depicts how the token ids of queries and responses are concatenated, from which the corresponding \textit{attention\_masks} and \textit{position\_ids} are derived and stored in the trajectory, as they are needed to recompute the logprobs of the updated policy model during Step IV.
Another crucial target is calculating multiple rule-based rewards based on the response texts.
These rule-based rewards, along with their summed scores, are also stored in the trajectory. 
Finally, response length, an important factor in evaluating reasoning capability \citep{guo2025deepseekR1}, is recorded in the trajectory. See~\cref{sec:evaluation_scheme} for detailed evaluation metrics.

\paragraph{Step IV: Policy Update}

Once trajectories required for updates are prepared, the first is to estimate the token-level KL divergence between current policy and reference model, scaled by a coefficient $\beta_{rew}$ as the KL reward.  
The summed scores, which are then appended to the last valid position (i.e., \texttt{<EOS>}) of the KL reward as total rewards.  
Next, following the iterative formula in~\cref{eq:ppo_loss}, total rewards are accumulated token by token in a recursive manner to estimate advantages. %
The policy logprobs are updated during each parameter update. These probabilities are calculated in chunks, with the chunk size potentially differing from that used in Step III. Consequently, the vision input must be re-collected and re-processed, which is key to ensuring the correct flow of visual data throughout the pipeline.
The updated policy logprobs, along with the old logprobs stored in trajectories, are used to compute the clipped ratio for policy loss calculation, as shown in~\cref{eq:ppo_loss}.
Besides, the KL divergence between the current policy and reference model is then estimated and weighted by a coefficient $\beta_{\text{loss}}$ to compute the KL loss.  
Finally, the total loss is computed using~\cref{eq:ppo_loss}, and policy parameters are updated. 
In total, updates are performed \textit{N} = \((\text{batch\_size} \,//\, \text{ppo\_batch\_size}) \times \text{ppo\_epochs}\) times.  
At this point, a single iteration of VLM-RL training is completed. The process is then repeated across all four parts while observing key metrics and evaluating performance.
\section{\titlename Scheme: Tracking Training Dynamics in RL for LLMs/VLMs}
\label{sec:evaluation_scheme}
\begin{figure}[tb!]
    \centering
    \includegraphics[width=0.98\textwidth]{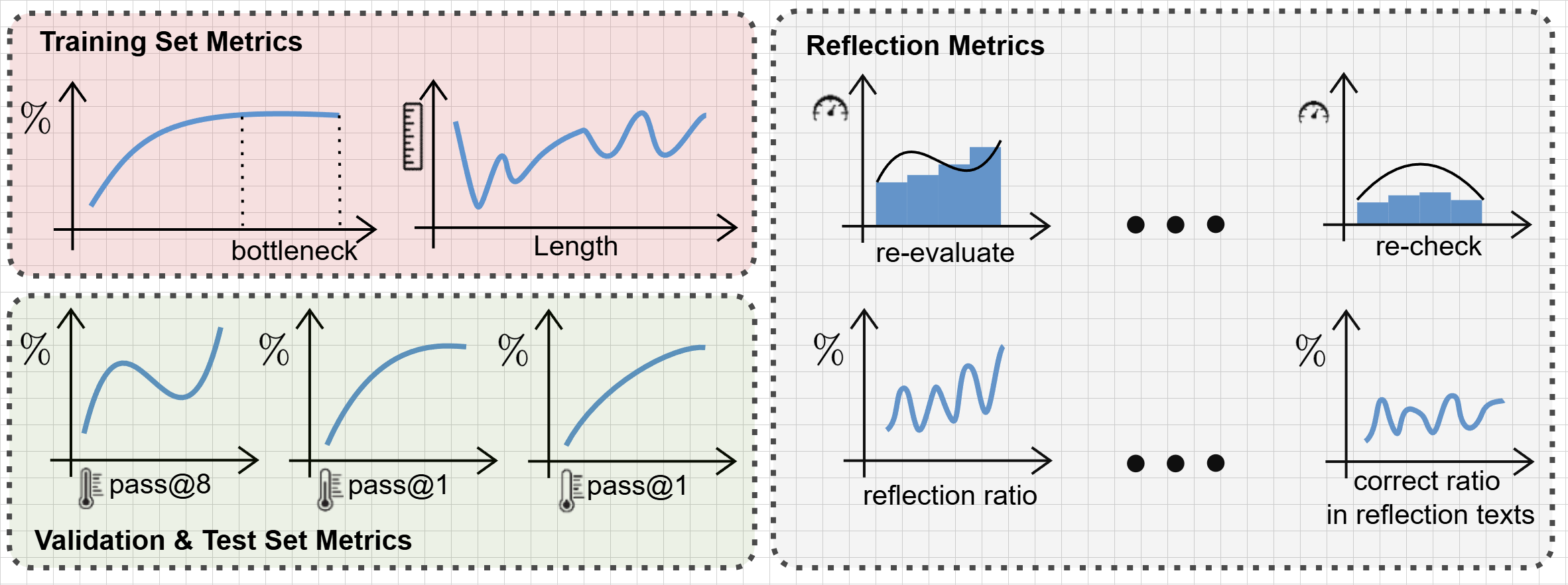}
\caption{Overview of evaluation metrics.}
    \label{fig:evaluation_scheme}
\end{figure}

Reliable evaluation has long been a challenge in RL research~\citep{agarwal2021deep}.  
Despite the growth of RL-based post-training for LLMs/VLMs, a unified and standardized evaluation scheme remains lacking.  
Here outlines the evaluation scheme used in the experiments, as shown in~\cref{fig:evaluation_scheme}.  
It categorizes evaluation metrics into three aspects: \textit{Train Set Metrics}, \textit{Validation/Test Set Metrics}, and \textit{Reflection Metrics}, aiming to establish a more rigorous and reliable assessment scheme.

\paragraph{General settings}
In RL evaluation, learning curves are commonly used to visualize training dynamics, with the y-axis representing key metrics such as cumulative rewards or accuracy.
The x-axis often represents two types of steps: \textit{generation steps} and \textit{gradient steps}, with \textit{generation steps} being preferred for clearer sample efficiency measurement and allow for fairer comparisons, as response generation typically takes longer than gradient updates.
Here, for accuracy learning curves, we advocate using \textit{epochs} as the x-axis label for improved interpretability, facilitating comparisons akin to those in SFT, where progress is tracked over dataset passes.

Additionally, due to the inherent fragility of RL algorithms
~\citep{henderson2018deep, andrychowicz2020matters}, factors such as different random seeds and initialization states can significantly impact training outcomes~\citep{colas2018many}. In traditional RL research, multiple runs (e.g., five, ten, or even dozens) are typically conducted, with the mean and error bars reported in learning curves to ensure statistical reliability.
In the context of LLMs/VLMs training, to balance computational cost and result stability, the mean learning curve from three independent runs should be reported.

\subsection{Training Set Metrics}
\paragraph{Accuracy curves} 
Training set accuracy reflects the correctness and effectiveness of both the algorithm and data preparation. Accuracy is recorded cumulatively per batch and logged per epoch. The main purpose is to illustrate training dynamics, while true performance should be assessed on the validation and test sets.
A typical training accuracy curve initially rises and then stabilizes. The stabilization phase, or \textit{bottleneck period}, indicates convergence and helps decide when to halt training.
Ideally, evaluation should include accuracy up to the bottleneck period for a comprehensive understanding of training dynamics.

\paragraph{Response length}
It reflects the model’s output pattern, including its level of detail and reasoning depth, can be shaped by RL training. 
Empirical results (~\cref{sec:training_results}) show that as responses become longer, models exhibit more reflective behaviors, contributing to improved generalization \citep{chu2025RLgeneral}.
Hence, response length serves as a crucial metric for monitoring the training process.

\subsection{Validation \& Test Set Metrics}
\paragraph{Accuracy curves}
Evaluation on the validation and test sets is critical for accurately assessing the model's capability and generalization.
Therefore, accurate accuracy measurements are essential, with online evaluation for small datasets and offline evaluation for larger ones.

Three sets of inference parameters are used to provide a comprehensive view of the model's performance: 1) pass@8, temperature=1.0, top\_p=1.0; 2) pass@1, temperature=0.6, top\_p=1.0; 3) pass@1, temperature=0.01, top\_p=0.001.
The first set evaluates the model's upper bound, while the second and third assess true performance, with the second preventing endless repetitions or incoherent outputs~\citep{deepseek_r1}, and the third following the VLM benchmark setting \citep{bai2023Qwen-VL}.
In practice, longer CoT models benefit from setting 2), while shorter response models are better reflected by setting 3).
These three settings ensure a balanced assessment of the model, highlighting both its maximum potential and true capabilities.
\paragraph{Accuracy tabs}
In addition to using curves to dynamically visualize and compare performance, static numerical tables are required to provide a clear summary of performance changes.
Since accuracy fluctuates throughout the training process, both the mean and maximum accuracy over all epochs are reported. These values are averaged across multiple runs to ensure statistical reliability.

\subsection{Reflection Metrics}
\paragraph{Words count}
Reflective behavior (or "aha moments") in models signals the effectiveness of RL training. However, the challenge lies in designing a mechanism to observe changes in this behavior over time.
Tracking the frequency of reflective words directly measures the model's reflective reasoning, revealing patterns in self-correction and problem-solving strategies.
A curated list of 15 reflective words: [`\textit{``re-check}'', ``\textit{re-evaluate}'', ``\textit{re-examine}'', ``\textit{re-think}'', ``\textit{recheck}'', ``\textit{reevaluate}'', ``\textit{reexamine}'', ``\textit{reevaluation}'', ``\textit{rethink}'', ``\textit{check again}'', ``\textit{think again}'', ``\textit{try again}'', ``\textit{verify}'', ``\textit{wait}'', ``\textit{yet}''] is tracked by counting their frequency during each \textit{generation\_steps}, as inspired by \cite{deepscaler2025} and \cite{xie2025logicrlunleashingllmreasoning}.

\paragraph{Ratio curves}
\begin{wraptable}{r}{0.50\textwidth}
\centering
\resizebox{0.50\textwidth}{!}{
\vspace{-2mm}
\begin{tabular}{c|c}
\hline
\textbf{Ratio Name} & \textbf{Formula} \\
\hline
reflection\_ratio & $\frac{\mathcal{N}_{ref}}{\mathcal{N}}$ \\
reflection\_ratio\_in\_correct\_answers & $\frac{\mathcal{N}_{ref+}}{\mathcal{N}_{+}}$ \\
reflection\_ratio\_in\_incorrect\_answers & $\frac{\mathcal{N}_{ref} - \mathcal{N}_{ref+}}{\mathcal{N} - \mathcal{N}_{+}}$ \\
correct\_ratio\_in\_reflection\_texts & $\frac{\mathcal{N}_{ref+}}{\mathcal{N}_{ref}}$ \\
correct\_ratio\_in\_no\_reflection\_texts & $\frac{\mathcal{N}_{+} - \mathcal{N}_{ref+}}{\mathcal{N} - \mathcal{N}_{ref}}$ \\
\hline
\end{tabular}
}
\caption{Definition of reflection ratios.}
\label{tab:reflection_ratios}
\end{wraptable}
Simply tracking word frequency is insufficient; it is also essential to observe how the proportion of reflective behavior changes and whether it contributes to accuracy improvement. To achieve this, five ratio metrics are designed, and the corresponding formulas are provided in~\cref{tab:reflection_ratios}, where $\mathcal{N}$ is the number of responses per batch, $\mathcal{N}_{ref}$ is the number of responses with reflection words, $\mathcal{N}_{+}$ is the number of correct responses per batch, and $\mathcal{N}_{ref+}$ is the number of correct responses with reflection words. These metrics quantify different aspects of reflection: the overall proportion of reflective responses, their distribution among correct and incorrect answers, and the accuracy differences between responses with and without reflection.
\section{Experiment}
\label{sec:experiment}
This section presents an evaluation of RL for VLMs, focusing on training and generalization aspects.
First, the correctness of the proposed framework is validated by evaluating performance across different VLMs and datasets, including \textit{mm\_math5k} \citep{sun2024mmmath} and \textit{geometry3k} \citep{lu2021inter}.
Performance improvements on validation and test sets are measured, as discussed in~\cref{sec:valid_test_results}.
Second, key RL training metrics are analyzed according to the scheme in~\cref{sec:evaluation_scheme}, covering epoch-wise accuracy and insights into the relationship between response length, reflection word ratio, and aha moments.
Finally, RL’s generalization ability is assessed, especially in comparison to SFT on high-quality data (see ~\cref{sec:rl_sft_generalization}).
\begin{figure}[tb!]
    \centering
    \begin{subfigure}{0.49\textwidth}
        \centering
        \includegraphics[width=\linewidth]{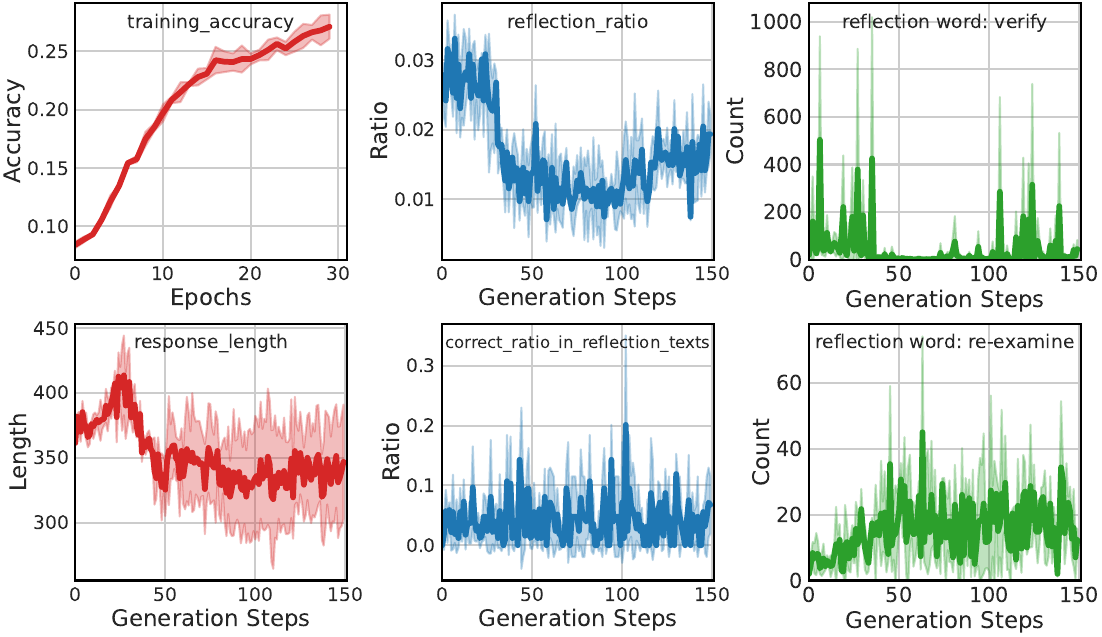}
        \caption{Qwen2-VL-Instruct-7B@\textit{mm\_math5k}}
        \label{fig:qwen2_mmmath_training}
    \end{subfigure}
    \begin{subfigure}{0.49\textwidth}
        \centering
        \includegraphics[width=\linewidth]{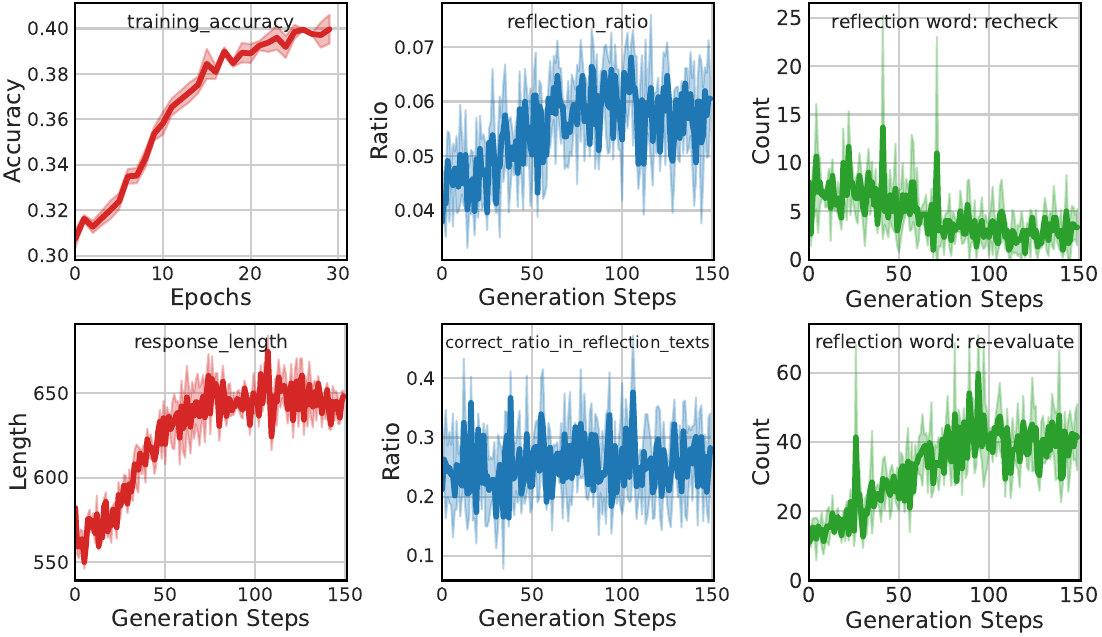}
        \caption{Qwen2.5-VL-Instruct-7B@\textit{mm\_math5k}}
        \label{fig:qwen2_5_mmmath_training}
    \end{subfigure}
    \begin{subfigure}{0.49\textwidth}
        \centering
        \includegraphics[width=\linewidth]{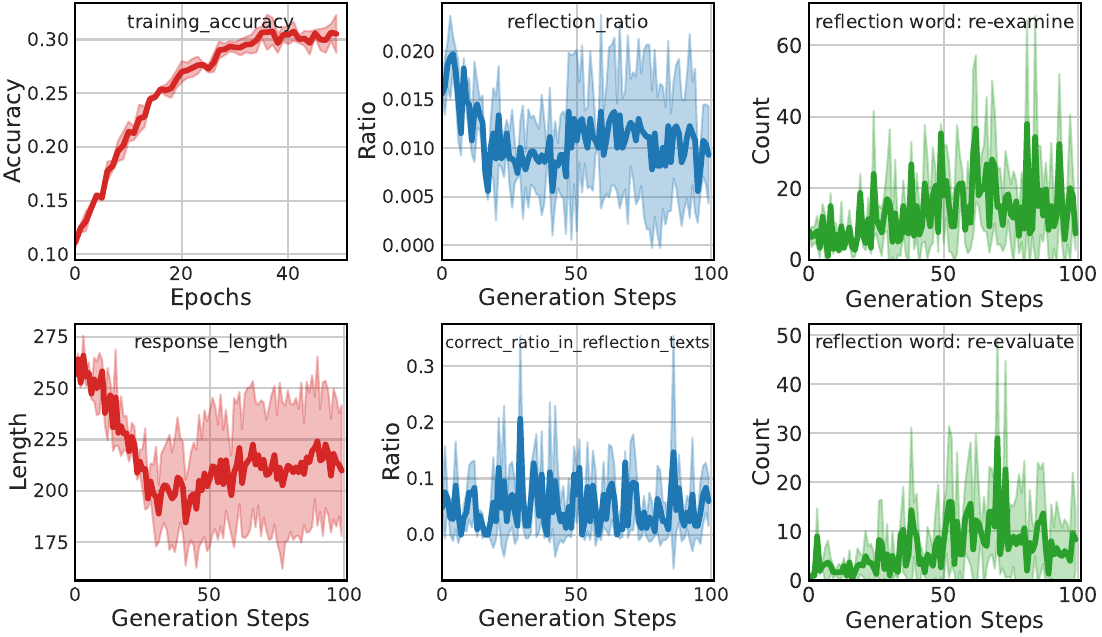}
        \caption{Qwen2-VL-Instruct-7B@\textit{geometry3k}}
        \label{fig:qwen2_geo3k_training}
    \end{subfigure}
    \begin{subfigure}{0.49\textwidth}
        \centering
        \includegraphics[width=\linewidth]{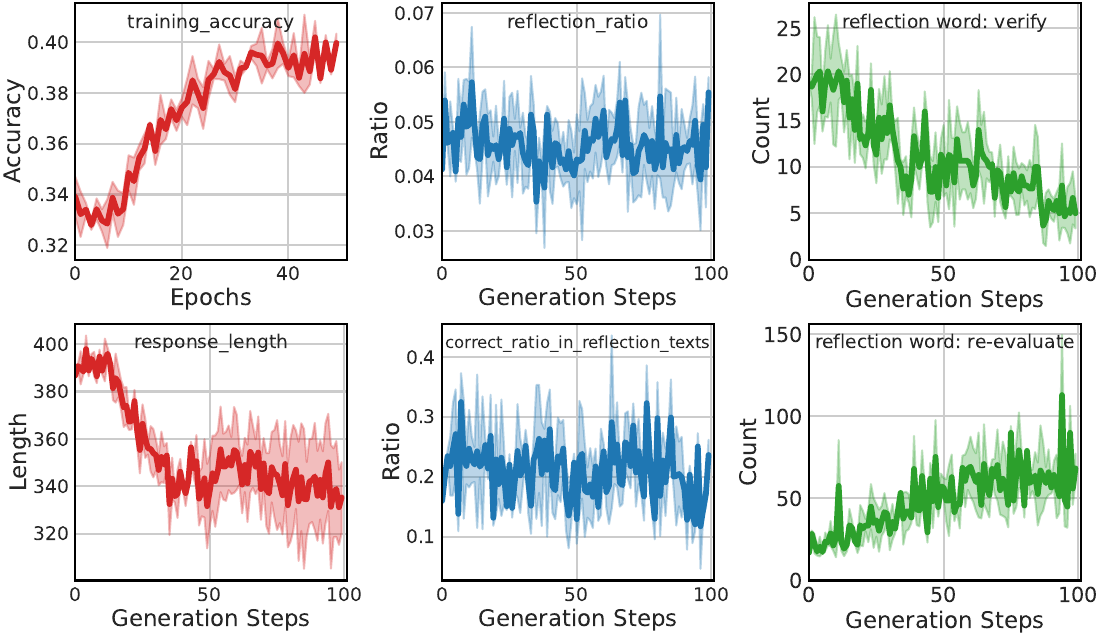}
        \caption{Qwen2.5-VL-Instruct-7B@\textit{geometry3k}}
        \label{fig:qwen2_5_geo3k_training}
    \end{subfigure}
    \caption{Training set metrics across models and datasets. 
    Red curves show training accuracy (per epoch) and response length (per generation step). 
    Blue curves depict key reflection ratios from~\cref{sec:evaluation_scheme}, and green curves illustrate the usage trends of the two most frequent and dynamic reflection words per experiment. 
    Shaded regions represent standard deviation across three runs.}
    \label{fig:training_metrics}
\end{figure}
\subsection{Setup}

\paragraph{Settings} In this work, only the LLM backend of VLM is trained, with the ViT encoder and connector frozen. For answer pattern extraction, the model is instructed to reason step by step, and the final answer is enclosed in \texttt{\textbackslash boxed{}}. Only accuracy and language rewards are applied, omitting format and token-level KL rewards.
Format reward is easily learned and may limit exploration space~\cite{zeng2025simplerlzooinvestigatingtamingzero}.
Token-level KL rewards are excluded to avoid reference model influence on advantage estimation, as recommended in \cite{xie2025logicrlunleashingllmreasoning}.
All experiments are conducted independently three times to ensure robustness, with the average of each evaluation metric reported across runs.

\paragraph{Parameters} The learning rate is set to $5.0\times10^{-6}$ with a warmup and cosine decay scheduler. \textit{Batch\_size} is $128$, and \textit{forward\_batch\_size} is $16$. Training is conducted for 1 \textit{ppo\_epochs} and batch is divided into $32$ minibatches, resulting in $32$ off-policy updates per batch. Generation settings include temperature and top\_p both set to $1.0$ and max length $2048$ tokens. All experiments are run on 8×H800 GPUs, with 7 allocated for training and 1 for inference. The total batch size for response collection is 896. The same hyperparameter settings are shared across experiments.
\textit{mm\_math5k} is trained for 30 epochs, corresponding to 150 generation steps, while \textit{geometry3k} is trained for 50 epochs, resulting in 100 generation steps.
\subsection{Training Set Results and Analysis}
\label{sec:training_results}
\cref{fig:training_metrics}
presents key training metrics across four experimental settings. The red lines represent the epoch-wise accuracy on the training set (top-left) and the response length trend over generation steps (bottom-left).
Training accuracy consistently increases, indicating that RL optimization is functioning as expected.
Response length serves as a useful diagnostic signal, reflecting the model’s generation pattern and output richness. Its variation is influenced by model architecture (see~\cref{fig:qwen2_mmmath_training,fig:qwen2_5_mmmath_training}), data distribution (see~\cref{fig:qwen2_5_mmmath_training,fig:qwen2_5_geo3k_training}), and even random seed (see the widening shaded area in late training stages).
Notably, a steady increase in response length is observed in Qwen2.5-VL-Instruct-7B trained on \textit{mm\_math5k}, suggesting that the model adopts a more elaborate reasoning style as training progresses.

\begin{figure}[tb!]
    \centering
    \begin{subfigure}{0.49\textwidth}
        \centering
        \includegraphics[width=\linewidth]{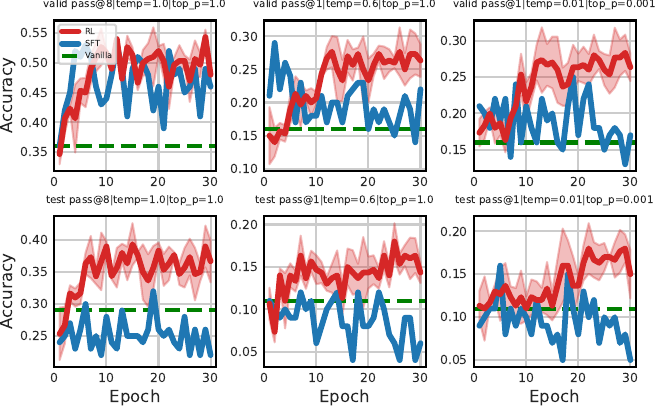}
        \caption{Qwen2-VL-Instruct-7B@\textit{mm\_math5k}}
        \label{fig:qwen2_sft_rl}
    \end{subfigure}
    \begin{subfigure}{0.49\textwidth}
        \centering
        \includegraphics[width=\linewidth]{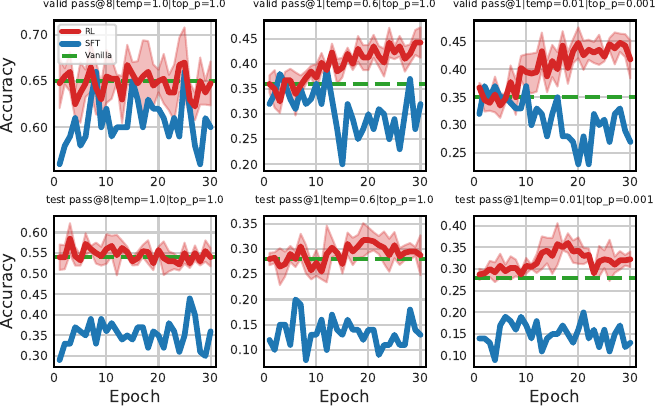}
        \caption{Qwen2.5-VL-Instruct-7B@\textit{mm\_math5k}}
        \label{fig:qwen2_5_sft_rl}
    \end{subfigure}
    \begin{subfigure}{0.49\textwidth}
        \centering
        \includegraphics[width=\linewidth]{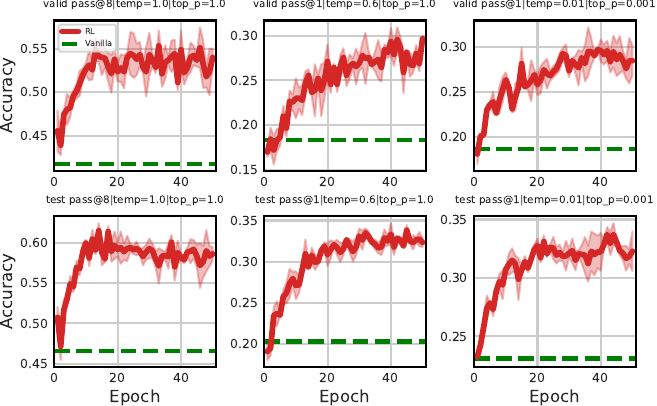}
        \caption{Qwen2-VL-Instruct-7B@\textit{geometry3k}}
        \label{fig:qwen2_geo3k}
    \end{subfigure}
    \begin{subfigure}{0.49\textwidth}
        \centering
        \includegraphics[width=\linewidth]{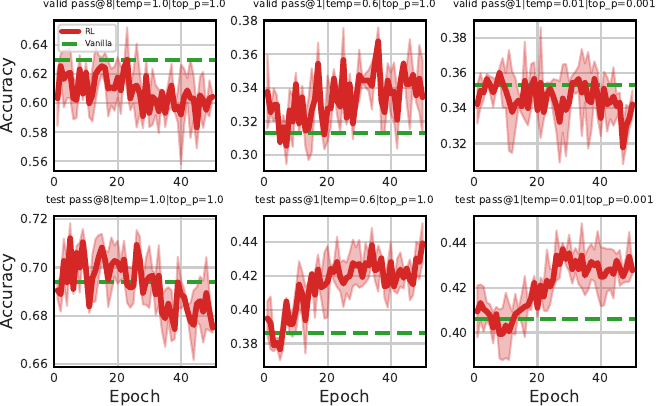}
        \caption{Qwen2.5-VL-Instruct-7B@\textit{geometry3k}}
        \label{fig:qwen2_5_rl_geo3k}
    \end{subfigure}
    \caption{Validation and test accuracy curves across training epochs for different VLMs and datasets. Red lines denote RL, blue lines denote SFT (see~\cref{sec:rl_sft_generalization}), and green indicate untrained (Vanilla) performance. All curves are averaged over 3 runs, with shaded areas indicating standard deviation.}
    \label{fig:rl_valid_test}
\end{figure}
\subsection{Reflection Metrics and Analysis}
\label{sec:reflection_results}
\cref{fig:training_metrics} presents key statistics on reflective behavior during training. The blue curves show \textit{reflection\_ratio} and \textit{correct\_ratio\_in\_reflection\_texts}, which capture how often reflection appears and whether it aids in correct reasoning. A full overview of all five ratios is in \cref{fig:reflection_ratios}.
The green curves show two representative reflection words per experiment, selected based on frequency and variation. Full trends are in \cref{fig:reflection_counts1,fig:reflection_counts2}.
Qwen2.5-VL consistently shows higher reflection and correct-in-reflection ratios than Qwen2-VL, suggesting reflective reasoning may be embedded in its pretraining corpus.
Still, reflection remains a minority behavior, and performance gains are primarily driven by improvements in non-reflective reasoning.
A key analytical focus is the relationship between \textit{response length}, \textit{reflection\_ratio}, and specific reflection words. Across all experiments, reflection ratio strongly correlates with response length, suggesting reflection contributes significantly to output length variation.
However, length and reflection variation do not always track accuracy. In (a) and (c), length decreases while accuracy improves; in (b), reflection ratio rises but correct reflection ratio remains stable (20–30\%).
In Qwen2-VL, \textit{verify} spikes early then fluctuates; in Qwen2.5-VL, richer expressions like \textit{re-evaluate} and \textit{re-examine} rise steadily, suggesting stylistic and behavioral differences.
In summary, while reflection and length reveal aspects of reasoning, performance remains the ultimate indicator.
\paragraph{``Aha Moments''}
\label{sec:valid_test_results}
An "aha moment" refers to the model’s ability to identify and correct its own reasoning errors during rollout \citep{guo2025deepseekR1}. 
As illustrated in~\cref{sec:aha_moments}, examples are provided in which different VLMs generate reflective reasoning chains that successfully lead to correct answers.
It is important to note that instances of such behavior can already be observed in base models~\citep{DrGRPO}. RL training amplifies this behavior, enhancing it rather than creating it from scratch.
Even after reflection, minor perceptual errors may persist, indicating that RL could further enhance perceptual grounding to improve overall model capacity.
While capturing “aha moments” is valuable, the main focus should be on improvements in validation and test accuracy, as discussed in the next section.
\begin{table}[tb!]
    \centering
    \renewcommand{\arraystretch}{1.2} 
    \begin{adjustbox}{max width=\textwidth}
    \setlength{\tabcolsep}{4pt}
    \begin{tabular}{c c c c c c c c c c c c c}
        \specialrule{1.0pt}{0pt}{0pt}
        \multirow{2}{*}{\textbf{Dataset}} & \multirow{2}{*}{\textbf{Model}} & \multirow{2}{*}{\shortstack{\textbf{Generation} \\ \textbf{Config}}} & \multicolumn{5}{c}{\textbf{Validation Set}} & \multicolumn{5}{c}{\textbf{Test Set}} \\
        \cmidrule(lr){4-8} \cmidrule(lr){9-13}
        & & & \textbf{Vanilla} & \textbf{SFT (Mean)} & \textbf{RL (Mean)} & \textbf{SFT (Max)} & \textbf{RL (Max)} & \textbf{Vanilla} & \textbf{SFT (Mean)} & \textbf{RL (Mean)} & \textbf{SFT (Max)} & \textbf{RL (Max)} \\
        \midrule
        \multirow{6}{*}{\textit{mm\_math5k}}
        & \multirow{3}{*}{\shortstack{Qwen2-VL-\\Instruct-7B}}
            & pass@8 temp=1.0  & 36.0 & \cellcolor[HTML]{F1B4B4}46.8 & \cellcolor[HTML]{EDA3A3}48.5 & \cellcolor[HTML]{E47171}53.0 & \cellcolor[HTML]{DE4949}55.7 & 29.0 & \cellcolor[HTML]{B7DFB7}25.2 & \cellcolor[HTML]{F4C7C7}35.2 & \cellcolor[HTML]{FADFDF}32.0 & \cellcolor[HTML]{E16060}42.7 \\
        &   & pass@1 temp=0.6  & 16.0 & \cellcolor[HTML]{F8DADA}19.9 & \cellcolor[HTML]{EFA9A9}23.4 & \cellcolor[HTML]{E47070}29.0 & \cellcolor[HTML]{DB3B3B}31.0 & 11.0 & \cellcolor[HTML]{C8E7C8}8.7  & \cellcolor[HTML]{F3C1C1}14.1 & \cellcolor[HTML]{FADFDF}12.0 & \cellcolor[HTML]{E36868}19.7 \\
        &   & ~~pass@1 temp=0.01 & 16.0 & \cellcolor[HTML]{F9E1E1}18.8 & \cellcolor[HTML]{EEA4A4}24.1 & \cellcolor[HTML]{EB9696}25.0 & \cellcolor[HTML]{DB3B3B}31.0 & 11.0 & \cellcolor[HTML]{C2E4C2}9.6  & \cellcolor[HTML]{F3C0C0}14.1 & \cellcolor[HTML]{EC9898}16.0 & \cellcolor[HTML]{E16060}20.7 \\
        \cmidrule(lr){2-13}
        & \multirow{3}{*}{\shortstack{Qwen2.5-VL-\\Instruct-7B}}
            & pass@8 temp=1.0  & 65.0 & \cellcolor[HTML]{D2ECD2}60.6 & \cellcolor[HTML]{FEF4F4}64.8 & \cellcolor[HTML]{FDF0F0}66.0 & \cellcolor[HTML]{F1B5B5}70.0 & 54.0 & \cellcolor[HTML]{64BC64}35.2 & \cellcolor[HTML]{FDF2F2}54.6 & \cellcolor[HTML]{8ECC8E}44.0 & \cellcolor[HTML]{E98989}61.8 \\
        &   & pass@1 temp=0.6  & 36.0 & \cellcolor[HTML]{BFE3BF}30.8 & \cellcolor[HTML]{F7D6D6}39.8 & \cellcolor[HTML]{F7D5D5}39.0 & \cellcolor[HTML]{E67A7A}47.2 & 28.0 & \cellcolor[HTML]{4CB14C}13.3 & \cellcolor[HTML]{FBE9E9}29.0 & \cellcolor[HTML]{B4DEB4}20.0 & \cellcolor[HTML]{EB9595}35.2 \\
        &   & ~~pass@1 temp=0.01 & 35.0 & \cellcolor[HTML]{BADFBA}31.2 & \cellcolor[HTML]{F3C1C1}40.3 & \cellcolor[HTML]{F8DADA}37.0 & \cellcolor[HTML]{E67A7A}46.2 & 28.0 & \cellcolor[HTML]{58B758}15.0 & \cellcolor[HTML]{F3C1C1}31.8 & \cellcolor[HTML]{A8D9A8}20.0 & \cellcolor[HTML]{E88484}38.3 \\
        \midrule
        \multirow{6}{*}{\textit{geometry3k}}
        & \multirow{3}{*}{\shortstack{Qwen2-VL-\\Instruct-7B}}
            & pass@8 temp=1.0  & 41.7 & - & \cellcolor[HTML]{E98989}52.5 & - & \cellcolor[HTML]{DE4949}57.0 & 46.6 & - & \cellcolor[HTML]{E98A8A}58.0 & - & \cellcolor[HTML]{DE4949}62.1 \\
        &   & pass@1 temp=0.6  & 18.3 & - & \cellcolor[HTML]{EFA9A9}25.2 & - & \cellcolor[HTML]{DB3B3B}30.3 & 20.3 & - & \cellcolor[HTML]{EA8C8C}30.3 & - & \cellcolor[HTML]{E16060}34.3 \\
        &   & ~~pass@1 temp=0.01 & 18.7 & - & \cellcolor[HTML]{EEA6A6}26.7 & - & \cellcolor[HTML]{DB3B3B}31.1 & 23.1 & - & \cellcolor[HTML]{E57676}31.1 & - & \cellcolor[HTML]{E05C5C}34.1 \\
        \cmidrule(lr){2-13}
        & \multirow{3}{*}{\shortstack{Qwen2.5-VL-\\Instruct-7B}}
            & pass@8 temp=1.0  & 63.0 & - & \cellcolor[HTML]{E5F5E5}60.8 & - & \cellcolor[HTML]{FDF1F1}64.4 & 69.4 & - & \cellcolor[HTML]{FEFCFC}69.3 & - & \cellcolor[HTML]{FDF0F0}71.8 \\
        &   & pass@1 temp=0.6  & 31.3 & - & \cellcolor[HTML]{FADFDF}33.4 & - & \cellcolor[HTML]{EC9999}37.9 & 38.6 & - & \cellcolor[HTML]{F7D6D6}41.5 & - & \cellcolor[HTML]{E98989}44.6 \\
        &   & ~~pass@1 temp=0.01 & 35.3 & - & \cellcolor[HTML]{F5FAF5}34.6 & - & \cellcolor[HTML]{FADFDF}37.7 & 40.6 & - & \cellcolor[HTML]{FBE5E5}42.0 & - & \cellcolor[HTML]{EC9898}45.0 \\
        \specialrule{1.0pt}{0pt}{0pt}
    \end{tabular}
    \end{adjustbox}
    \vspace{2mm}
    \caption{Mean and maximum accuracy on validation \& test sets averaged across 3 runs. RL consistently outperforms the untrained (Vanilla) baseline across all settings. Cell colors indicate relative improvement: deeper red denotes larger gains over Vanilla, while green indicates degradation.}
    \vspace{-2mm}
    \label{tab:results1}
\end{table}

\subsection{Validation \& Test set Results and Analysis}
\cref{fig:training_metrics} shows the accuracy dynamics, with red curves for RL-trained VLMs, blue for SFT (discussed in~\cref{sec:rl_sft_generalization}), and green dashed lines for the untrained (Vanilla) model.
Each curve shows the mean over 3 independent runs, with shaded regions indicating standard deviation.
\cref{tab:results1} summarizes the mean and maximum accuracy for all epochs on the validation and test sets across different generation settings.
Color intensity reflects improvement relative to Vanilla: darker red indicates higher gains, while green represents underperformance.

Notable performance improvements are observed on both validation and test sets. RL consistently yields significant gains across all generation settings.
On \textit{mm\_math5k}, RL achieves a 1.35× average increase in accuracy, peaking at 1.76×. Similarly, on geometry3k, RL brings an average gain of 1.36×, with a maximum of 1.51×.
Even for Qwen2.5-VL-Instruct-7B, already among the strongest VLMs of its size, RL continues to enhance generalization, improving pass@1 test accuracy on \textit{mm\_math5k} by 3.5\%, with a peak gain of 10\%. For geometry3k, RL improves by 1.4\%, up to 4.8\%.
These results demonstrate that RL can effectively enhance both in-distribution and out-of-distribution performance of strong vision-language models, even when baseline capabilities are already very high.

\subsection{Generalization on visual mathematical tasks: RL versus SFT}
\label{sec:rl_sft_generalization}
Since the \textit{mm\_math} dataset \citep{sun2024mmmath} provides CoT solutions from textbooks, these high-quality responses can serve as supervision signals.
A key objective is to compare the generalization ability of RL and SFT, a topic of ongoing debate in the research community \citep{chu2025RLgeneral, ye2025limo}.
SFT is performed on Qwen2/2.5-VL-Instruct-7B for the same number of epochs as RL, using the \textit{mm\_math5k} dataset with golden CoT solutions.
The learning rate follows a warm-up cosine decay schedule with an initial value of $1 \times 10^{-5}$, and the batch size is set to 16.
Performance is evaluated on the validation and test sets after each epoch, as shown in~\cref{fig:rl_valid_test}.

Our findings are summarized as follows:
1) RL outperforms SFT across all configurations and models, with the gap widening as training progresses.
2) On the test set (OOD queries), SFT occasionally underperforms the untrained baseline, indicating overfitting to the training distribution. In contrast, RL achieves higher accuracy than both SFT and the baseline, demonstrating stronger generalization.

In summary, the advantages of RL for VLMs are threefold: 1) It does not require high-quality responses, often scarce in multimodal scenarios \citep{guo2024mammothvl}. 2) Queries can be reused multiple times, improving sample efficiency. 3) RL maintains strong generalization in vision mathematical tasks, while SFT is limited by poor out-of-distribution performance.
\section{Related Work}
Recent efforts in RL for VLMs focus on enhancing reasoning for visual mathematics~\citep{meng2025mm, Huang2025VisionR1IR, Peng2025LMMR1E3, chen2025r1v} and extending RL to broader visual tasks such as grounding, detection, and classification~\citep{liu2025visual, shen2025vlmr1}.
While these works advance the frontier, this report addresses two foundational gaps: 1) the absence of a concise framework outlining RL training for VLMs, and 2) the lack of a structured evaluation framework tailored for RL training.
%
%
Unlike feature-rich RL toolkits like TRL~\citep{vonwerra2022trl}, verl~\citep{sheng2024hybridflow}, and OpenRLHF~\citep{hu2024openrlhf}, which prioritize performance and complexity, our framework offers a minimalist, from-scratch implementation focused on transparency and ease of customization, without competing on performance.
Evaluation practices for RL-based LLM/VLM training are still under-standardized, making comparison difficult.
This report introduces a unified evaluation scheme with metrics covering both performance and behavioral aspects of RL training.
A concurrent effort, SimpleRL-Zoo \citep{zeng2025simplerlzooinvestigatingtamingzero}, also highlights the importance of robust evaluation in LLMs under zero-settings.
Compared to this, this work offers finer-grained analysis of reflective behavior and more comprehensive tracking of accuracy dynamics.

\section{Conclusion and Future Work}
This work introduces a minimalist and reproducible RL framework for VLMs, built entirely from scratch, alongside a standardized evaluation scheme for tracking performance dynamics and reflective behaviors.
Empirical findings offer significant insights into the interplay between reflection, response length, and generalization, showing RL’s superior performance over SFT.
In future work, the framework will be further refined for improved usability, simplicity, and extensibility. Leveraging its modular and extensible design, we plan to explore its application to emerging architectures, such as VLMs with linear attention~\citep{minimax01}, and even extend RL scaling to fully autoregressive image generation settings~\citep{openai2025introducing}. Meanwhile, the evaluation scheme will be continuously enhanced to provide deeper and more comprehensive insights into model behavior across these diverse scenarios.

\bibliography{colm2025_conference}
\bibliographystyle{colm2025_conference}

\appendix
\clearpage
\newpage
\section{Hyper-Parameters}
\label{sec:parameters}
\begin{itemize}
  \item General training setup: These parameters control the core training loop, including the number of epochs and batch size. \\ \textit{batch\_size=128}; \textit{epochs=30(geometry3k), 50(mm\_math5k)}.

  \item Model component training configuration: Specifies which parts of the model are trainable. \\ \textit{train\_vit=False}; \textit{train\_connector=False}; \textit{train\_llm=True}

  \item Optimization and numerical precision: Sets gradient clipping and computation precision to ensure training stability and efficiency. \\
  \textit{clip\_grad\_norm=1.0}; \textit{dtype=bfloat16}

  \item PPO-related parameters: Define how policy optimization is performed, including the number of PPO passes, clipping thresholds, and reward normalization. \\ \textit{ppo\_epochs=1}; \textit{forward\_batch\_size=16}; \textit{ppo\_batch\_size=4}; \textit{ppo\_backward\_batch\_size=4}; \textit{gradient\_accumulation\_steps=1}, \textit{epsilon=0.2}, \textit{gamma=1.0}
  \item Reward shaping and regularization: These parameters control KL Loss penalties and KL reward modifications to balance exploration and stability. \\ \textit{kl\_loss\_coeff=0.001}, \textit{kl\_reward\_coeff=0.0}

  \item vLLM Inference and sampling configuration: Controls how outputs are generated during training, including sequence length and sampling strategy. \\\textit{max\_tokens=2048}; \textit{top\_p=1.0}; \textit{temperature=1.0}; \textit{gpu\_memory\_utilization=0.8}
\end{itemize}
\begin{table}[ht]
    \centering
    \renewcommand{\arraystretch}{1.2}
    \begin{tabular}{l p{8.5cm}}
        \hline
        Term & Definition \\
        \hline
        \textit{batch\_size} & Number of queries per GPU for response collection. \\
        \textit{forward\_batch\_size} & Number of query-responses processed in a single forward pass to obtain logits. Due to GPU memory constraints, only a subset of the sampled responses can be forwarded at a time. \\
        \textit{ppo\_batch\_size} & Size of mini-batches into which the sampled query-responses of \textit{batch\_size} are divided. It allows for a degree of off-policy updates, facilitated by PPO-clip loss. \\
        \textit{ppo\_backward\_batch\_size} & Number of query-responses processed per backward pass. This value is computed as ppo\_batch\_size // gradient\_accumulation\_steps. \\
        \textit{epochs} & Number of iterations over RL queries, which is consistent with the concept of data epochs in SFT. \\  
        \textit{ppo\_epochs} & The number of times a batch of query-response pairs is reused. A higher number of updates indicates a greater degree of off-policy learning. \\  
        \textit{generation\_steps} & Number of generating iterations, where each call to \texttt{llm.generate} increments the count by one. \\
        \textit{gradient\_steps} & Number of gradient backward steps, incremented by one with each call to \texttt{loss.backward}. \\
        \hline
    \end{tabular}
    \caption{Definitions of Batch and Step-related Terms}
    \vspace{-5mm}
    \label{tab:batch_step_terms}
\end{table}
\clearpage
\section{Reflection Ratios}
\begin{figure}[ht!]
    \centering
    \begin{subfigure}{0.49\textwidth}
        \centering
        \includegraphics[width=\linewidth]{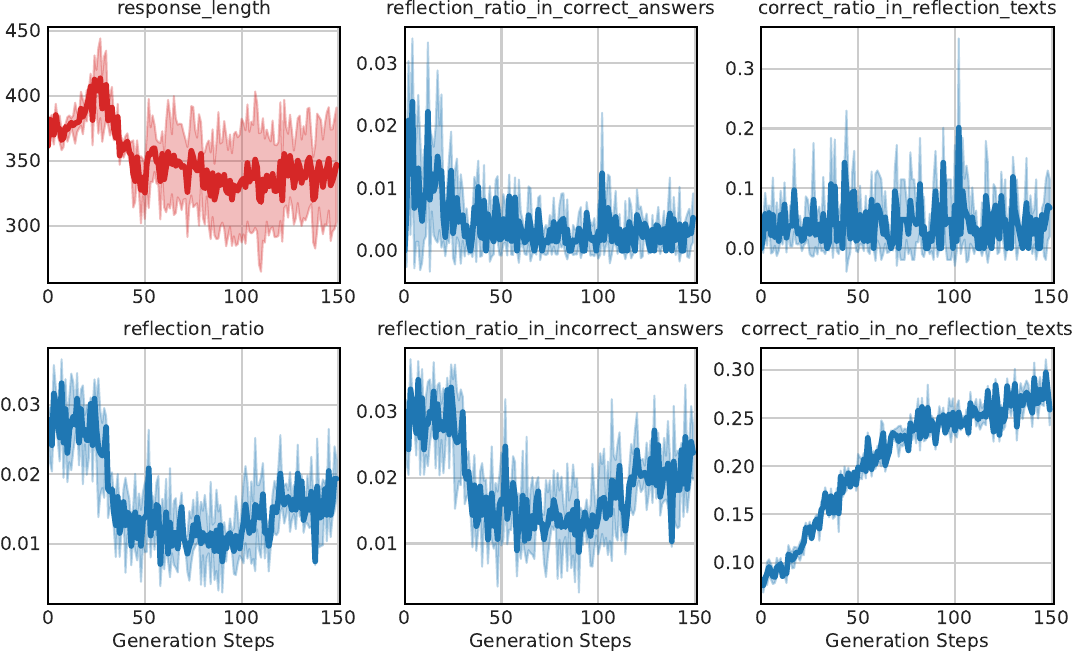}
        \caption{Qwen2-VL-Instruct-7B@\textit{mm\_math5k}}
        \label{fig:qwen2_mmmath_reflection}
    \end{subfigure}
    \begin{subfigure}{0.49\textwidth}
        \centering
        \includegraphics[width=\linewidth]{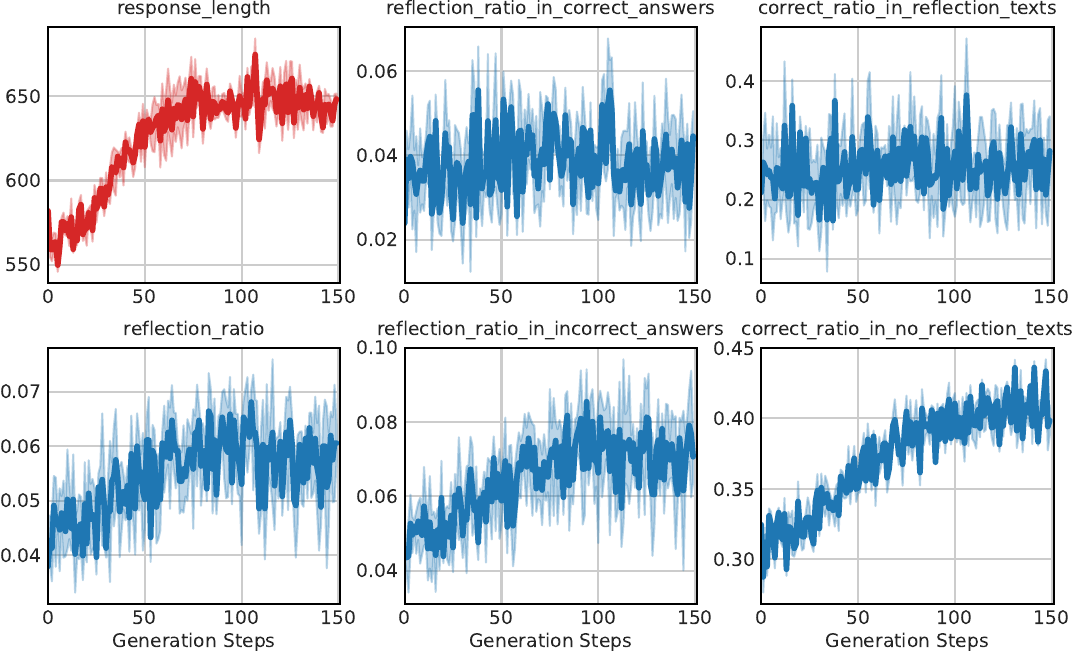}
        \caption{Qwen2.5-VL-Instruct-7B@\textit{mm\_math5k}}
        \label{fig:qwen2_5_mmmath_reflection}
    \end{subfigure}
    \begin{subfigure}{0.49\textwidth}
        \centering
        \includegraphics[width=\linewidth]{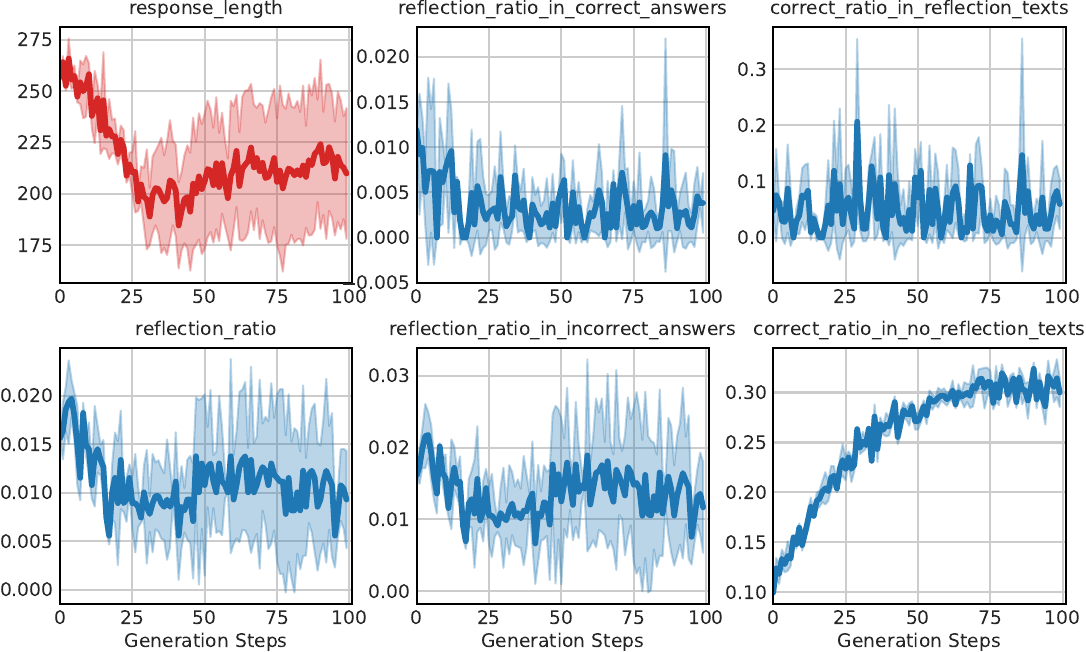}
        \caption{Qwen2-VL-Instruct-7B@\textit{geometry3k}}
        \label{fig:qwen2_geo3k_reflection}
    \end{subfigure}
    \begin{subfigure}{0.49\textwidth}
        \centering
        \includegraphics[width=\linewidth]{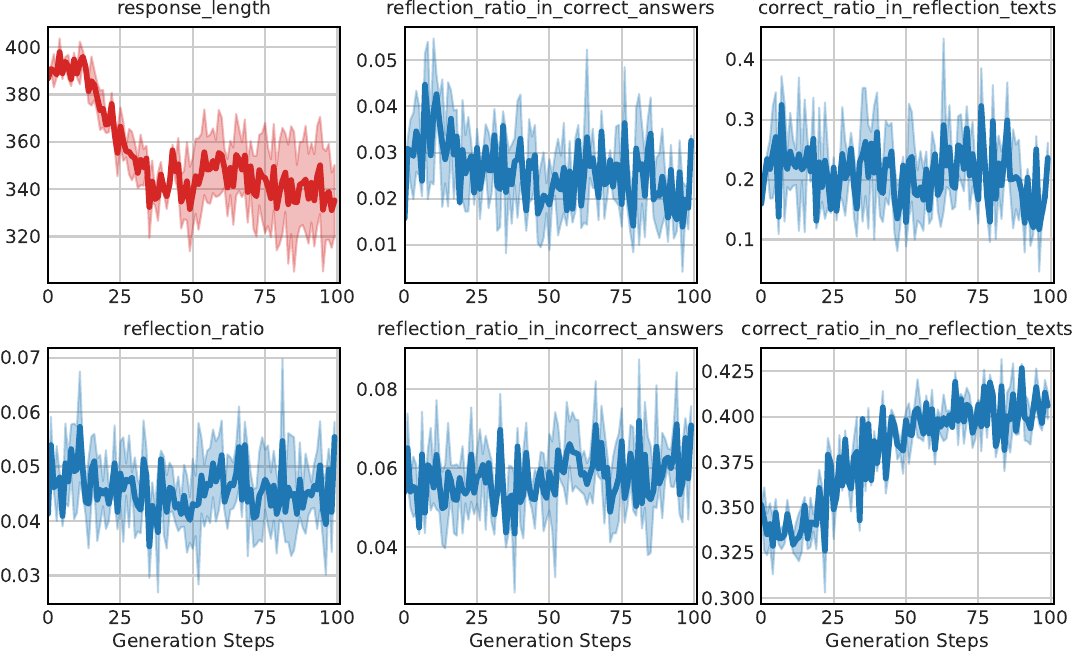}
        \caption{Qwen2.5-VL-Instruct-7B@\textit{geometry3k}}
        \label{fig:qwen2_5_geo3k_reflection}
    \end{subfigure}
    \caption{Reflection Ratios}
    \label{fig:reflection_ratios}
\end{figure}
\clearpage

\section{Reflection Word Counts}
\begin{figure}[h!]
    \centering
    \begin{subfigure}{0.90\textwidth}
        \centering
        \includegraphics[width=\linewidth]{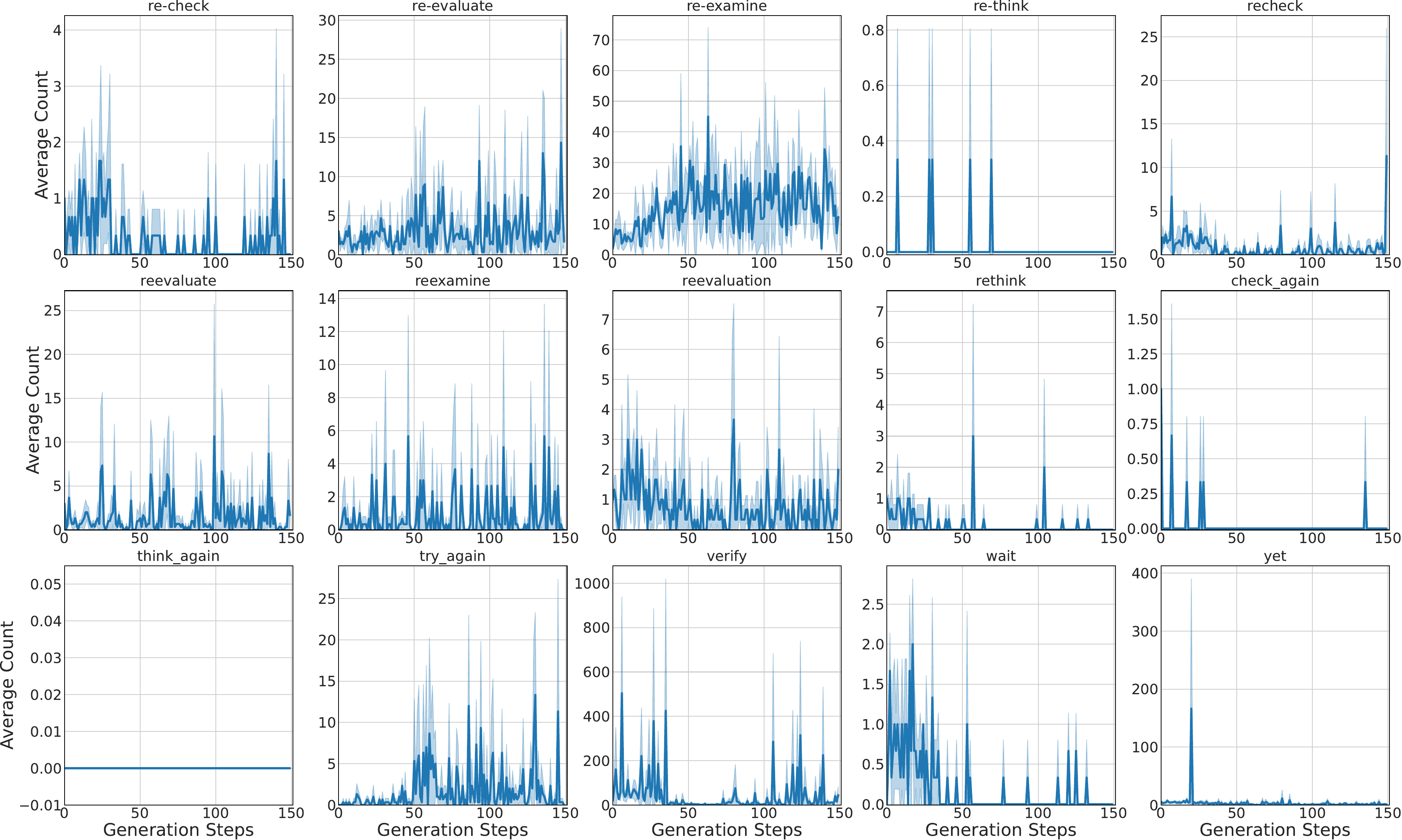}
        \caption{Qwen2-VL-Instruct-7B@\textit{mm\_math5k}}
        \label{fig:qwen2_mmmath_counts}
    \end{subfigure}
    \begin{subfigure}{0.90\textwidth}
        \centering
        \includegraphics[width=\linewidth]{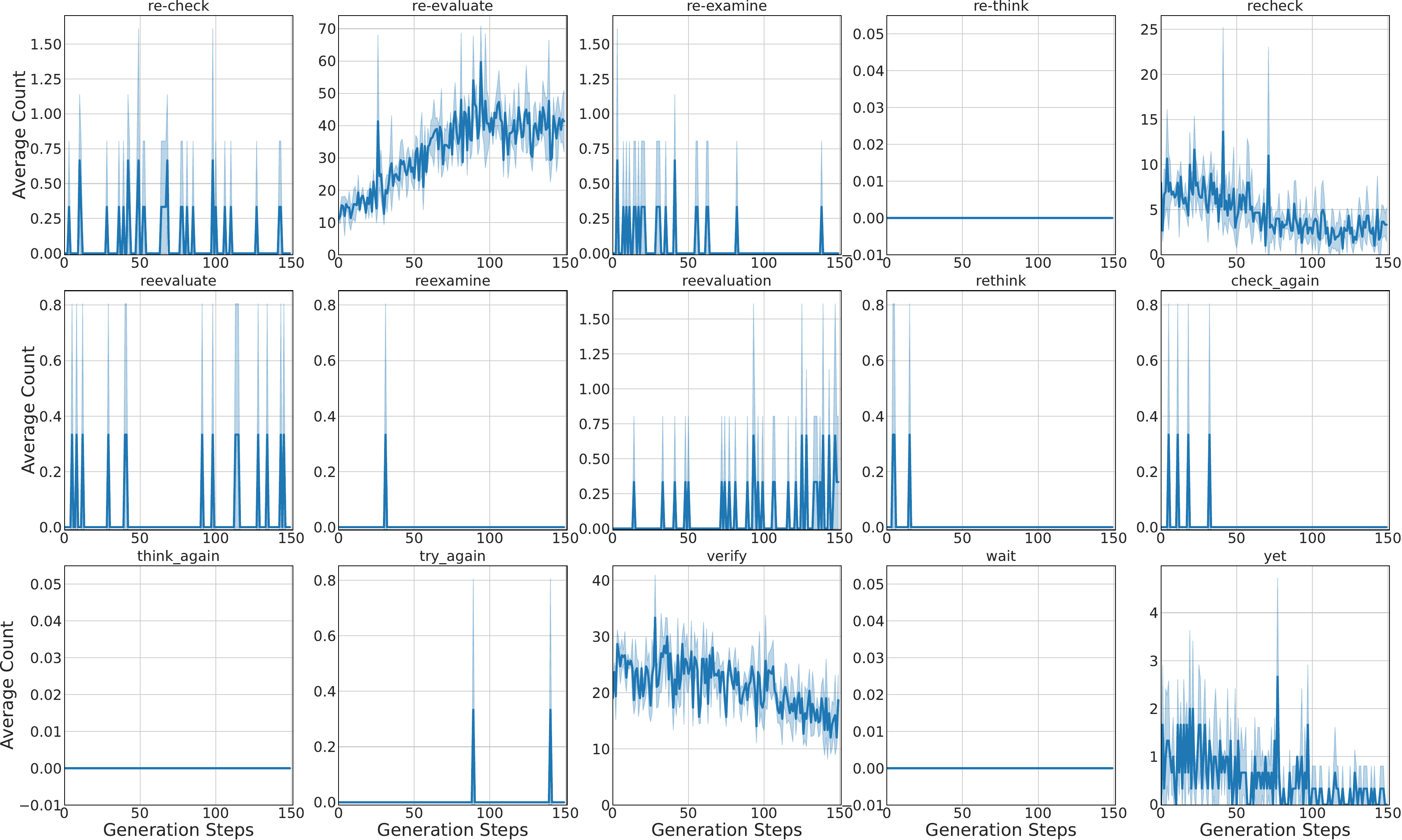}
        \caption{Qwen2.5-VL-Instruct-7B@\textit{mm\_math5k}}
        \label{fig:qwen2_5_mmmath_counts}
    \end{subfigure}
    \caption{Reflection Counts}
    \label{fig:reflection_counts1}
\end{figure}

\begin{figure}[tb!]
    \centering
    \begin{subfigure}{0.90\textwidth}
        \centering
        \includegraphics[width=\linewidth]{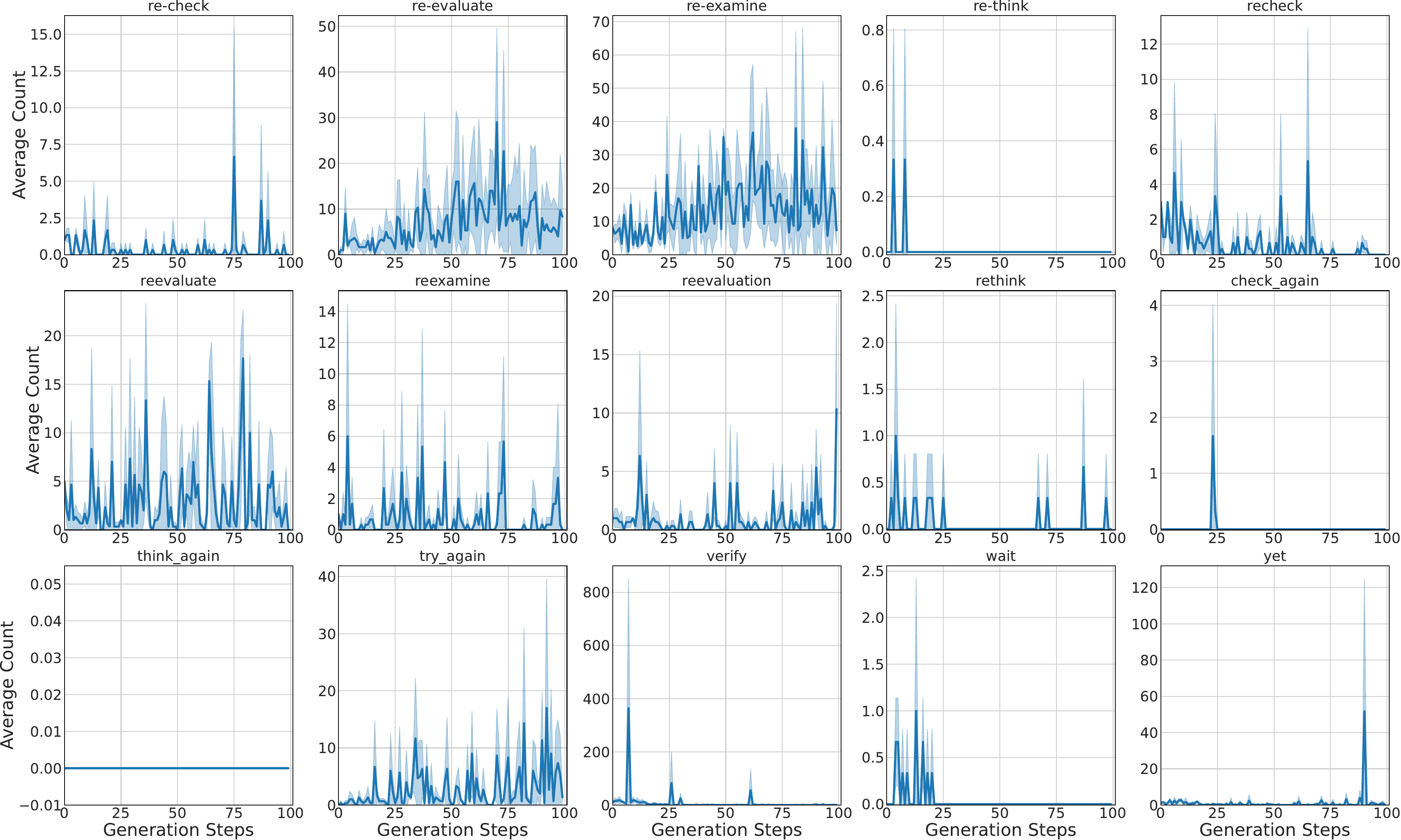}
        \caption{Qwen2-VL-Instruct-7B@\textit{geometry3k}}
        \label{fig:qwen2_geo3k_counts}
    \end{subfigure}
    \begin{subfigure}{0.90\textwidth}
        \centering
        \includegraphics[width=\linewidth]{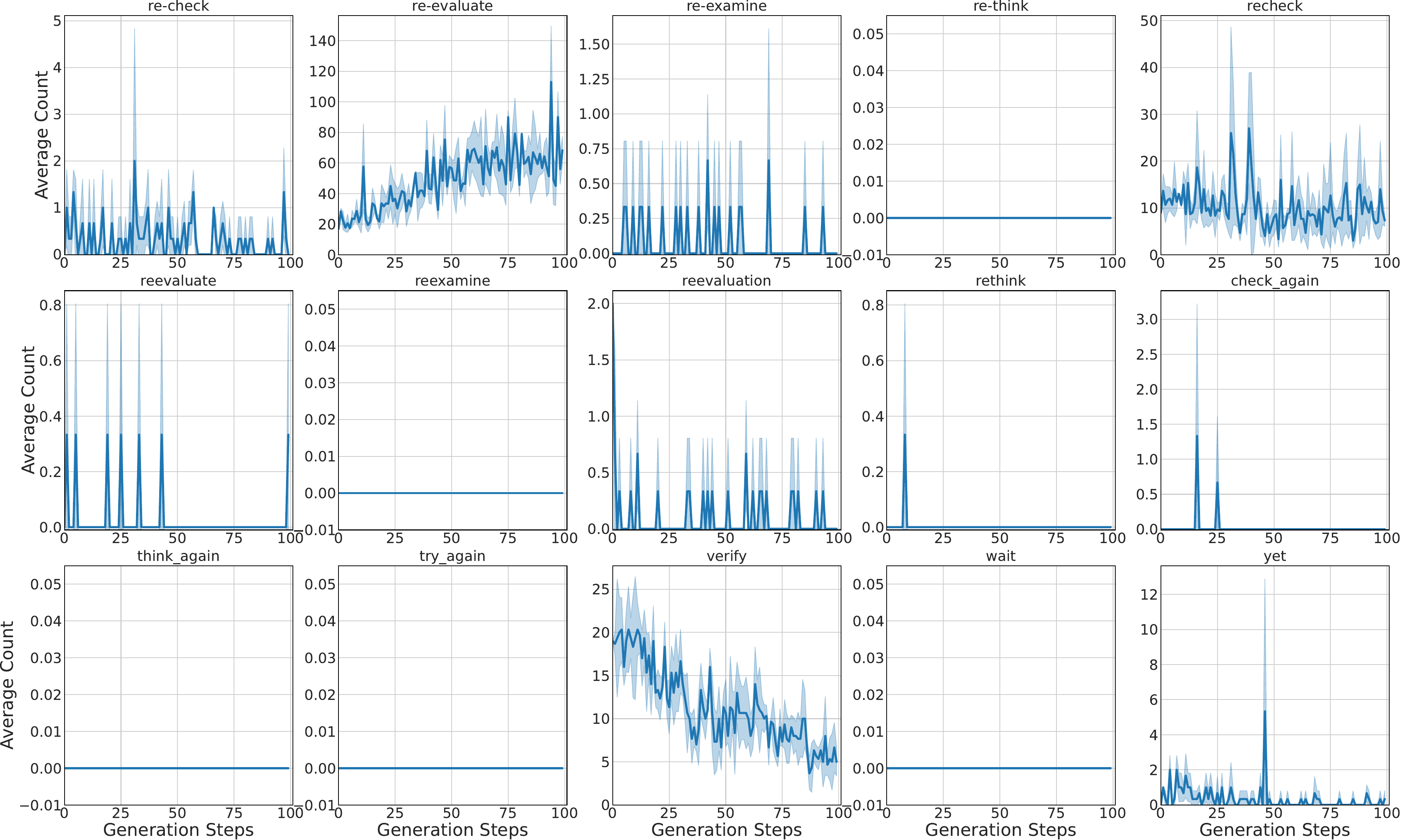}
        \caption{Qwen2.5-VL-Instruct-7B@\textit{geometry3k}}
        \label{fig:qwen2_5_geo3k_counts}
    \end{subfigure}
    \caption{Reflection Counts}
    \label{fig:reflection_counts2}
\end{figure}
\clearpage

\clearpage

\section{``Aha Moments''}
\label{sec:aha_moments}
\begin{tcolorbox}[
  colback=blue!5, 
  colframe=blue!70!black, 
  boxrule=0.8pt, 
  sharp corners, 
  enhanced, 
  width=\textwidth, 
  title={\textbf{Qwen2.5-VL-Instruct 7B@mm\_math5k}},
  coltitle=white,
  fonttitle=\bfseries
]
  \begin{tabular}{@{}p{0.30\textwidth} p{0.69\textwidth}@{}}
    \begin{minipage}[t]{\linewidth}
      \vspace{0pt} 
      \includegraphics[width=\linewidth]{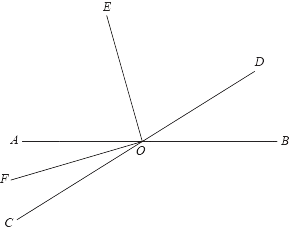}
    \end{minipage}
    &
    \begin{minipage}[t]{\linewidth}
      \vspace{25pt} 
      \textbf{Question:} As shown in Figure~(1), given point \(O\) is the intersection point of lines \(AB\) and \(CD\), 
      \(\angle BOD = 32^\circ\), \(OE\) bisects \(\angle AOD\), \(\angle EOF = 90^\circ\),
      find the degree measure of \(\angle COF\).
    \end{minipage}
  \end{tabular}
  
  \vspace{1.5em}
  \noindent
  \textbf{Response:} To solve the problem, we need to determine the measure of \(\angle COF\) given the information in the diagram and the conditions provided.
  
First, let's analyze the given information:
- \(\angle BOD = 32^\circ\)
- \(OE\) bisects \(\angle AOD\)
- \(\angle EOF = 90^\circ\).
Since \(OE\) bisects \(\angle AOD\), it means that \(\angle AOE = \angle DOE\). Let's denote the measure of \(\angle AOE\) and \(\angle DOE\) as \(x\). Therefore, we have:
\[
\angle AOD = \angle AOE + \angle DOE = x + x = 2x
\]
Next, we know that the angles around point \(O\) on a straight line sum up to \(180^\circ\). Since \(\angle AOD\) and \(\angle BOD\) are supplementary, we can write:
\[
\angle AOD + \angle BOD = 180^\circ
\]
Substituting the known values:
\[
2x + 32^\circ = 180^\circ
\]
Solving for \(x\):
\[
2x = 180^\circ - 32^\circ
\]
\[
2x = 148^\circ
\]
\[
x = 74^\circ
\]
Thus, \(\angle AOE = 74^\circ\) and \(\angle DOE = 74^\circ\).
Since \(\angle EOF = 90^\circ\), we can find \(\angle COF\) by subtracting \(\angle EOF\) from \(\angle AOE\):
\[
\angle COF = \angle AOE - \angle EOF
\]
\[
\angle COF = 74^\circ - 90^\circ
\]
\[
\angle COF = -16^\circ
\]
{\color{purple}\textbf{However, since angles cannot be negative, we need to re-evaluate the problem.}} The correct interpretation is that \(\angle COF\) is the complementary angle to \(\angle EOF\) within the context of the problem, which is:
\[
\angle COF = 180^\circ - \angle EOF - \angle DOE
\]
\[
\angle COF = 180^\circ - 90^\circ - 74^\circ
\]
\[
\angle COF = 16^\circ
\]
Thus, the correct measure of \(\angle COF\) is:
\[
\boxed{\color{blue}16}
\]
\end{tcolorbox}

\begin{tcolorbox}[
  colback=blue!5, 
  colframe=blue!70!black, 
  boxrule=0.8pt, 
  sharp corners, 
  enhanced, 
  width=\textwidth, 
  title={\textbf{Qwen2-VL-Instruct 7B@mm\_math5k}},
  coltitle=white,
  fonttitle=\bfseries
]
  \begin{tabular}{@{}p{0.30\textwidth} p{0.69\textwidth}@{}}
    \begin{minipage}[t]{\linewidth}
      \vspace{0pt}
      \includegraphics[width=\linewidth]{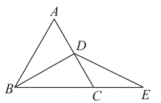}
    \end{minipage}
    &
    \begin{minipage}[t]{\linewidth}
    \vspace{15pt}
    \textbf{Question:} As shown in the figure, it is known that $\triangle ABC$ is an equilateral triangle, $BD$ is the median of $\triangle ABC$. Extend $BC$ to $E$ so that $CE=CD$, and connect $DE$. What is the measure of $\angle BDE$?
    \end{minipage}
  \end{tabular}
  \vspace{1.5em}
  \noindent
  \textbf{Response:} Since $$\triangle ABC$$ is an equilateral triangle, we have $$AB=BC=AC$$ and $\angle ABC=\angle BAC=\angle BCA=60^\circ$.
Since $$BD$$ is the median of $\triangle ABC$, we have $AD=DC$.
Therefore, $$\triangle ABD$$ and $$\triangle DCE$$ are congruent by SAS congruence.
This means that $\angle ABD=\angle DCE$.
Since $\angle ABC=\angle BAC=60^\circ$, we have $\angle BDE=\angle ABC-\angle ABD=\angle BAC-\angle DCE=60^\circ-60^\circ=0^\circ$.
However, this is not possible since $$\angle BDE$$ is an angle.
{\color{purple}\textbf{We made a mistake in our reasoning.
Let's try again.}}
Since $$\triangle ABD$$ and $$\triangle DCE$$ are congruent, we have $\angle ABD=\angle DCE$.
Therefore, $\angle BDE=\angle ABD+\angle BAC=60^\circ+60^\circ=120^\circ$.
So the measure of $$\angle BDE$$ is $\boxed{\color{blue}120^\circ}$.
\end{tcolorbox}

\begin{tcolorbox}[
  colback=red!5!white, 
  colframe=red!70!black, 
  boxrule=0.8pt, 
  sharp corners, 
  enhanced, 
  width=\textwidth, 
  title={\textbf{Qwen2.5-VL-Instruct 7B@geometry3k}},
  coltitle=white,
  fonttitle=\bfseries
]
  \begin{tabular}{@{}p{0.30\textwidth} p{0.69\textwidth}@{}}
    \begin{minipage}[t]{\linewidth}
      \vspace{0pt}
      \includegraphics[width=\linewidth]{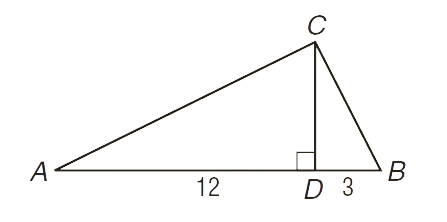}
    \end{minipage}
    &
    \begin{minipage}[t]{\linewidth}
    \vspace{15pt}
    \textbf{Question:} In $\triangle ABC$, $CD$ is an altitude and $\angle ACB=90^{\circ}$ If $AD=12$ and $BD=3,$ find $AC$ to the nearest tenth.
    \end{minipage}
  \end{tabular}
  
  \vspace{1.5em}
  \noindent
\textbf{Response:} To solve the problem, we need to find the length of $AC$ in $\triangle ABC$ where $CD$ is an altitude and $\angle ACB = 90^\circ$. We are given that $AD = 12$ and $BD = 3$. Since $D$ is the foot of the perpendicular from $C$ to $AB$, $CD$ is perpendicular to $AB$, making $\triangle ACD$ and $\triangle BCD$ right triangles.
We can use the Pythagorean theorem in $\triangle ACD$ and $\triangle BCD$ to find the length of $AC$.
In $\triangle ACD$, we have:
\[
AC^2 = AD^2 + CD^2
\]
\[
AC^2 = 12^2 + CD^2
\]
\[
AC^2 = 144 + CD^2
\]
In $\triangle BCD$, we have:
\[
BC^2 = BD^2 + CD^2
\]
\[
BC^2 = 3^2 + CD^2
\]
\[
BC^2 = 9 + CD^2
\]
Since $AC = AB$, we can set the two expressions for $AC^2$ equal to each other:
\[
144 + CD^2 = 9 + CD^2
\]
Subtract $CD^2$ from both sides:
\[
144 = 9
\]
{\color{purple}\textbf{This is a contradiction, so we need to re-evaluate the problem.}} Instead, we use the fact that $AC^2 = AD \cdot AB$ because $CD$ is the altitude to the hypotenuse of a right triangle, and the Pythagorean theorem in $\triangle ACD$ and $\triangle BCD$ confirms this relationship.
Calculate $AB$:
\[
AB = AD + BD = 12 + 3 = 15
\]
Now, using the relationship $AC^2 = AD \cdot AB$:
\[
AC^2 = 12 \cdot 15 = 180
\]
\[
AC = \sqrt{180} = \sqrt{36 \cdot 5} = 6\sqrt{5} \approx 13.4
\]
Thus, the length of $AC$ to the nearest tenth is:
\[
\boxed{\color{red}13.4}
\]
\end{tcolorbox}

\begin{tcolorbox}[
  colback=red!5!white, 
  colframe=red!70!black, 
  boxrule=0.8pt, 
  sharp corners, 
  enhanced, 
  width=\textwidth, 
  title={\textbf{Qwen2-VL-Instruct 7B@geometry3k}},
  coltitle=white,
  fonttitle=\bfseries
]
  \begin{tabular}{@{}p{0.60\textwidth} p{0.29\textwidth}@{}}
    \begin{minipage}[t]{\linewidth}
      \vspace{0pt} 
      \includegraphics[width=\linewidth]{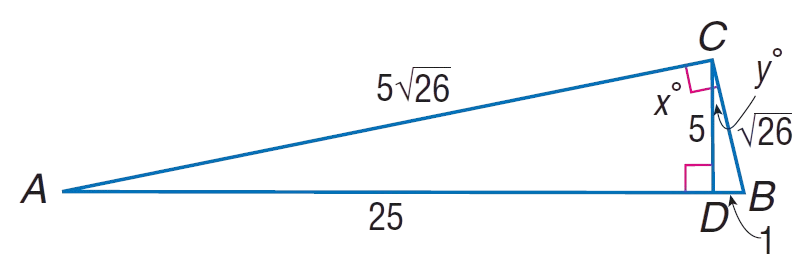}
    \end{minipage}
    &
    \begin{minipage}[t]{\linewidth}
    \vspace{35pt} 
    \textbf{Question:} Find \(\sin A\).
    \end{minipage}
  \end{tabular}
  
  \vspace{1.5em}
  \noindent
\textbf{Response:} To find \(\sin A\), we first need to determine the length of side \(AC\). We can use the Pythagorean theorem in right triangle \(ABC\), where \(AB = 25\), \(BC = 5\), and \(AC = 5\sqrt{26}\).
\[AB^2 + BC^2 = AC^2\]
\[25^2 + 5^2 = (5\sqrt{26})^2\]
\[625 + 25 = 250\]
\[650 = 250\]
{\color{purple}\textbf{This equation is not correct, indicating that there might be a mistake in the given lengths or angles. Let's re-examine the given information and the calculations.}}
Given that \(AB = 25\), \(BC = 5\), and \(AC = 5\sqrt{26}\), we can use the sine function to find \(\sin A\). The sine of an angle in a right triangle is the ratio of the length of the side opposite the angle to the length of the hypotenuse.
\[\sin A = \frac{BC}{AC}\]
\[\sin A = \frac{5}{5\sqrt{26}}\]
\[\sin A = \frac{1}{\sqrt{26}}\]
\[\sin A = \frac{\sqrt{26}}{26}\]
Therefore, the value of \(\sin A\) is \boxed{\color{red}\frac{\sqrt{26}}{26}}.
\end{tcolorbox}
\end{document}